\pdfoutput=1
\documentclass[11pt]{article}

\usepackage[preprint]{acl}

\usepackage{times}
\usepackage{latexsym}

\usepackage[T1]{fontenc}
\usepackage[utf8]{inputenc}

\usepackage{microtype}

\usepackage{inconsolata}

\usepackage{graphicx}

\usepackage{booktabs}
\usepackage{comment}
\usepackage{subcaption}
\usepackage{multirow}
\usepackage{graphicx}
\usepackage{caption}
\usepackage{amssymb}
\title{GMoE: Empowering LLMs Fine-Tuning via MoE Graph Collaboration}

\author{Ting Bai \\
  Beijing University of \\Posts and Telecommunication \\
  \texttt{baiting@bupt.edu.cn} \\\And
  Yue Yu \\
  Beijing University of \\Posts and Telecommunication\\
  \texttt{loadingyy@bupt.edu.cn} \\
  \And
  Le Huang \\
  Beijing University of \\Posts and Telecommunication\\
  \texttt{lehuang@bupt.edu.cn} \\
  \AND
  Zenan Xu \\
  \texttt{xuzn@alumni.sysu.edu.cn} \\\And
  Chuan Shi \\
  Beijing University of \\Posts and Telecommunication\\
  \texttt{shichuan@bupt.edu.cn} \\
  }

\begin{document}
\maketitle
\begin{abstract}
The sparse Mixture-of-Experts (MoE) architecture of large language models (LLMs) confronts an inherent issue of load imbalance arising from the simplistic linear router strategy, which ultimately causes the instability and inefficient learning of LLMs.
To address this challenge, we introduce a novel MoE graph-based framework \textbf{GMoE}, aimed at enhancing the collaboration among multiple experts.
In GMoE, a graph router function is designed to capture the collaboration signals among experts. This enables all experts to dynamically allocate information derived from input data by sharing information with their neighboring experts.
Moreover, we put forward two coordination strategies in GMoE: the \emph{Poisson distribution-based distinction strategy} and the \emph{Normal distribution-based balance strategy}, to further release the capacity of each expert and increase the model stability in the fine-tuning of LLMs. 
Specifically, we leverage a parameter-efficient fine-tuning technique, i.e., Low-Rank Adaptation (LoRA), to implement the graph MoE architecture.
Extensive experiments on four real-world benchmark datasets demonstrate the effectiveness of GMoE, showing the benefits of facilitating collaborations of multiple experts in LLM fine-tuning. The code of experimental implementation is available at \url{https://github.com/BAI-LAB/GMoE}.

% The sparse Mixture-of-Experts (MoE) architecture of large language models (LLMs) confronts an inherent issue of load imbalance arising from the simplistic linear router strategy, which ultimately causes the instability and inefficient learning of LLMs. To address this challenge, we introduce a novel MoE graph-based framework GMoE, aimed at enhancing the collaboration among multiple experts. In GMoE, a graph router function is designed to capture the collaboration signals among experts. This enables all experts to dynamically allocate information derived from input data by sharing information with their neighboring experts. Moreover, we put forward two coordination strategies in GMoE: the Poisson distribution-based distinction strategy and the Normal distribution-based balance strategy, to further release the capacity of each expert and increase the model stability in the fine-tuning of LLMs. Specifically, we leverage a parameter-efficient fine-tuning technique, i.e., Low-Rank Adaptation (LoRA), to implement the graph MoE architecture. Extensive experiments on four real-world benchmark datasets demonstrate the effectiveness of GMoE, showing the benefits of facilitating collaborations of multiple experts in LLM fine-tuning.

%The code of experimental implementation is available at \url{https://github.com/BAI-LAB/GMoE}.
\end{abstract}

\section{Introduction}
The Mixture-of-Expert (MoE) architecture has emerged as a promising approach to enhance the overall performance of large language models (LLMs) under the scaling law theory. The application of MoE in the fine-tuning of LLMs has drawn significant attention due to the substantial improvements it brings to model performance in practical downstream applications.
In a typical MoE setup, a simple linear function, acting as the “router function” or “gating function,” assigns weights to each expert based on the information from input tokens, then the top-k experts with largest assigned weights are activated in the fine-tuning process. 
However, sparse MoE methods face significant challenges that impede their performance and introduce model instability, with a core issue being their linear router strategy that creates severe load imbalance, where a few experts are over-trained while others are under-utilized~\cite{switch-trans}.

Recent studies have introduced load balance losses to constrain expert allocation frequencies, thereby alleviating MoE load imbalance~\citep{switch-trans,loadbal,moe20222,moeloracontra,loramoe4,wang2022adamix}.
However, these approaches primarily focus on regularization-based constraints for load balancing while failing to address a critical limitation: the absence of effective communication and collaborative mechanisms among experts during router allocation. Without expert collaboration, the router relies solely on individual input patterns instead of leveraging collective capabilities. This coordination gap exacerbates inefficiencies: even with balanced allocations, the lack of inter-expert communication prevents dynamic optimization to exploit each expert’s unique strengths.
As a result, experts remain underutilized, limiting the model’s potential and leading to instability during LLM fine-tuning.

These challenges highlight a critical research gap: while load regularization alleviates imbalance, enhancing the stability and efficiency of LLM fine-tuning with MoE requires moving beyond standalone constraints to design architectures that enable effective collaboration among experts.
To address this issue, we propose \textbf{GMoE}, a novel graph-based MoE framework that explicitly models collaboration among experts. Specifically, GMoE introduces a graph router to encode collaborative signals among experts via a MoE Graph integrating input tokens and expert nodes. Using graph neural networks (GNNs), the router aggregates information iteratively, capturing both token features and inter-expert interactions. This enables experts to jointly process inputs and dynamically share knowledge, leading to more informed, collaborative routing decisions.

%GMoE introduces a graph router to encode collaborative signals among experts, operating on a MoE Graph that integrates input token nodes and expert nodes. 
% Leveraging graph neural networks (GNNs),
% the graph router captures both input token information and inter-expert interactions through iterative information aggregation and propagation. This allows experts to jointly understand input information while dynamically sharing knowledge with their neighborhoods, enabling more informed and collaborative routing decisions.
To further empower the capability of each expert and enhance their collaboration, we propose a novel coordination strategy: the \emph{Poisson distribution-based distinction strategy}. 
Specifically, the Poisson distribution loss function promotes specialization by encouraging experts to handle distinct input aspects, thereby stimulating their unique capabilities. Concurrently, the Normal distribution-based balance strategy regulates activation frequencies, naturally balancing the overall workload distribution.
This synergistic combination enables experts to discriminatively process inputs while maintaining workload balance, fostering coordinated interaction and enhancing overall model efficiency.
% the Poisson distribution loss function plays a crucial role in maintaining the distinctiveness of each expert's input-processing capabilities. It encourages experts to specialize in different aspects of the input information, release the different capabilities of each expert. The normal distribution is adopted to keep the balance of the activated frequency of each expert in a natural pattern.  
% This combination of Poisson distinction and normal balance distribution strategies strengthens experts' collaboration in our graph-based MoE framework, enabling experts to process inputs discriminatively while maintaining balanced activation frequencies that foster coordinated interaction.

We adopt the parameter-efficient fine-tuning (PEFT) technique, i,e.,  Low-Rank Adaptation (LoRA), to enable an efficient implementation of our graph MoE architecture for LLM fine-tuning. Through the expert collaboration mechanism in the graph router and two coordination strategies, our GMoE framework achieves state-of-the-art performance and remarkable stability during the fine-tuning of diverse benchmark datasets. Our contributions are summarized as follows:

\begin{itemize}
\item We propose a novel GNNs-based 
MoE framework GMoE to address the instability problem of MoE in LLMs fine-tuning. 
A graph router assigns weights to experts with consideration of their collaborative interactions on MoE Graph. 

\item We propose a novel coordination strategies, i.e., the Poisson distribution-based distinction strategy. Along with the Normal distribution-based load balance strategy, our graph based MoE architecture, GMoE, fully unleash the capabilities of individual experts and significantly enhance their collaboration.

\item Extensive experiments conducted on four real-world benchmark datasets with three typical base LLMs demonstrate the effectiveness of our GMoE in terms of model \textbf{Accuracy}, and \textbf{Stability},  showing the benefits of empowering collaborations of multiple experts of LLMs. 
\end{itemize}

\section{Related Work}

\subsection{Mixture-of-Expert (MoE)}
Empowered by the collaboration of multiple experts, the Mixture-of-Expert (MoE) framework has achieved remarkable performance in many applications~\citep{chen2024llava,li2024mixlora,lin2024moe,li2024uni,zadouri2023pushing}.
% such as recommendation, video understanding, natural language generation, and so on. 
Each expert in the MoE framework specializes in handling a subset of input information and different experts coordinate together to gain benefits for the overall MoE framework.
According to the number of experts activated by the router function, the MoE methods can be classified into two categories, i.e., dense MoE and sparse MoE~\citep{loramoe,li2024locmoe}.
In dense MoE, all experts are activated in the learning process, which enhances the model capability but suffers from high computational costs. 
In sparse MoE, only a selected subset of experts are activated to use their specialized knowledge and achieve optimal results, thereby reducing computational overhead.
% leverages their specialized knowledge for optimal results, which reduces the computational overhead but also leverages their specialized knowledge for optimal results. 
The participation of each expert in the computation process is assigned by a router function (or gating function), ensuring an optimal blend of their specialized contributions~\citep{moe20222,tang2025graphmoe}.
Specifically, the router is usually a simple linear function that controls the engagement of expert computations.
In the dense MoE, all experts are allocated to computation depending on the weights assigned by the router. In sparse MoE, only Top-K experts with the highest weights are selectively activated in the learning process~\citep{loadbal}. 

For example, the typical sparse MoE approach Switch Transformers~\citep{switch-trans} routes only a single expert for the input token to reduce the computation costs, but it faces the load imbalance problem due to lacking collaboration of multiple experts. 
% Existing studies add auxiliary losses~\citep{bengio2015conditional,switch-trans} to keep the load of each expert balance, but it still an open challenge to explore better ways to keep trade-off of the activated number of experts and load balance in the sparse MoE architecture. 
Existing studies have incorporated auxiliary losses~\citep{bengio2015conditional,switch-trans} to maintain a balanced load among experts in sparse MoE architectures. However, exploring an optimal collaboration mechanism among the activated experts to enhance their load distribution remains an ongoing challenge in this field.
In our work, we propose a novel graph router to assign weights to experts with consideration of their collaborative interactions on the MoE Graph, so as to achieve better capability and stability in fine-tuning of LLMs.  
\subsection{PEFT with MoE}
The aim of fine-tuning LLMs is to optimize the model's performance in specific downstream tasks. Due to the high computational costs of updating all parameters in the full fine-tuning approach, the Parameter-Efficient Fine-Tuning (PEFT) technique is proposed to address this problem~\citep{lora,houlsby2019parameter,li2021prefix,lester2021power,tian2024hydralora}.  In PEFT, only a small subset of parameters are updated of the base LLMs, making it more efficient and practical in real applications. 
% The PEFT methods can be generally classified into three types, i.e., LoRA~\citep{lora}, adapter tuning~\citep{houlsby2019parameter}, and prompt tuning~\citep{lester2021power}.
% Adapter tuning uses feed-forward up and feed-forward down projection matrices in the transformer block and only updates two adapter metrics in the fine-tuning process. Prompt tuning introduces a set of learnable prompt embeddings that are appended to the input tokens.
The most widely used PEFT technique is LoRA~\citep{lora}, which uses a low-rank decomposition technique and updates the decomposed parameter matrix in the model training. 

Recent studies have shown improvements of model performance by integrating the MoE framework with PEFT in LLMs fine-tuning literature~\citep{wu2024mixture,zadouri2023pushing}.
In the transformer block of LLMs, each expert is a LoRA module working on either the FFN layer~\citep{loramoe,wang2022adamix,li2024mixlora}, attention layer~\citep{moeloracontra,loramoe2,zhu2023sira}, the whole transformer block~\citep{zadouri2023pushing,loramoe3,mola} and each layer~\citep{wu2024mixture}.
Then a router function of MoE is designed to blend the contributions of all experts. 
Among them, the adoption of LoRA in the FFN layer attracted the most attention due to its extensive applicability in various LLM tasks in recent years. 

% Because per-taining LLMs requires massive datasets and a high demand for computational resources, 
Our work focuses on exploring the effective MoE collaboration mechanism to enhance the PEFT of LLMs in downstream tasks. 
% Our work aims to explore the effective MoE collaboration mechanism in the parameter-efficient fine-tuning of LLMs area. 
The MoE consists of multiple LoRA components that work on the FFN layer in the transformer block. 
Different from the conventional router function employed in existing MoE studies, where the weight assigned to each expert is solely dependent on its own input information,
z
we design a novel graph router, which leverages graph neural networks to learn the collaboration information among all experts based on a MoE graph.

    \begin{figure*}
        \centering
        \includegraphics[width=\textwidth]{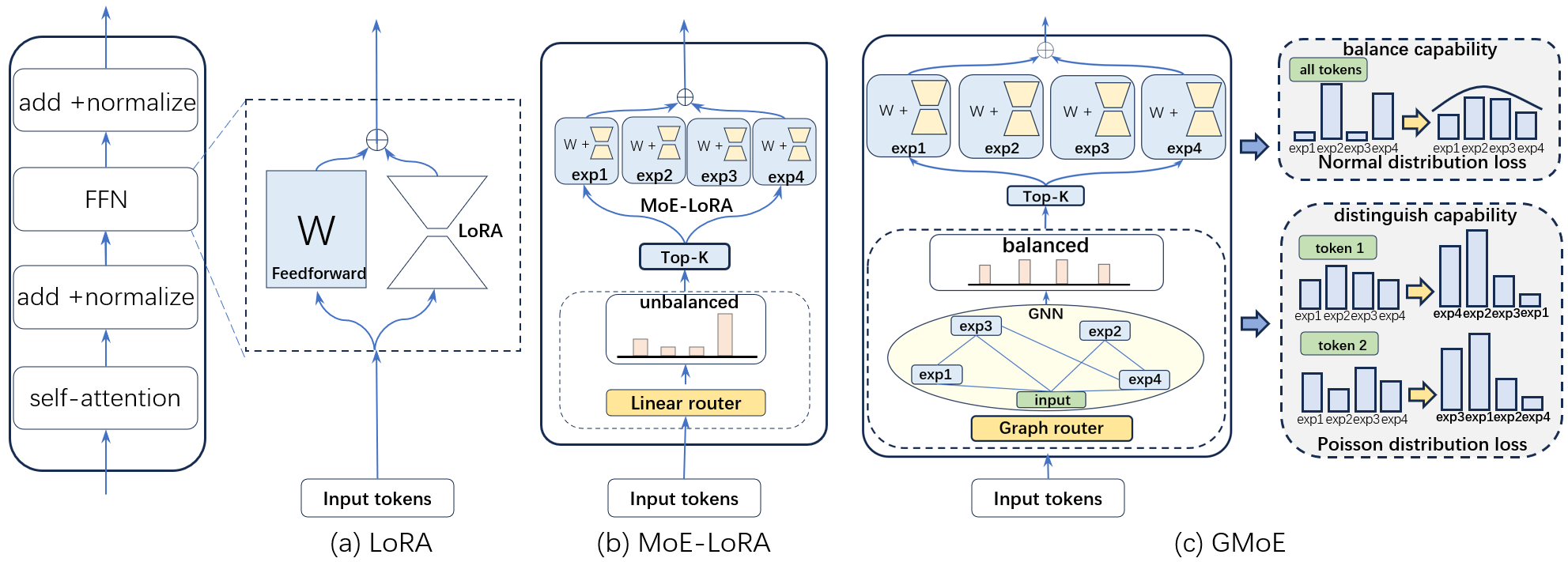}
        \caption{The overview of MoE architectures in the FFN layer. (a) The LoRA component is applied in the FFN layer of the transformer block. (b) The typical MoE architecture with LoRA in LLMs with a linear router function to assign weights. (c) Our proposed GMoE architecture with a graph router based on the MoE graph. For different input information (input1 and input2), the distinctive capability of experts is optimized by the Poisson distinction loss. For all input information, the activated frequency of each expert is balanced by the normal distribution loss.}
        \label{fig:moes}
    \end{figure*}

\section{Preliminary} 
\subsection{Low-Rank Adaptation (LoRA)}
Low-rank adaptation (LoRA)~\citep{lora} is a widely used parameter-efficient fine-tuning (PEFT) technique in pre-trained LLMs. LoRA adopts a low-rank decomposition technique to update the parameters of the decomposed parameter matrix to learn the data distribution of the specific downstream task.
It works on the feed-forward layer (FFN layer) in a transformer block of LLMs. For a linear layer in FFN represented as $\mathbf{h}=\mathbf{W}\mathbf{x}$, the update of the decomposed parameter matrix of LoRA is defined as:
\begin{equation}
\mathbf{h} = \mathbf{W}\mathbf{x} + \Delta \mathbf{W} \mathbf{x} = \mathbf{W}\mathbf{x} + \frac{\alpha}{r} \mathbf{B}\mathbf{A}\mathbf{x},
\label{equ:lora}
\end{equation}
where $\mathbf{x} \in \mathbb{R}^{I}$ is the representation of input information, and $\mathbf{W} \in \mathbb{R}^{{O}\times{I}}$ is the pre-trained parameter matrix of LLMs. $\mathbf{A} \in \mathbb{R}^{r \times I}$ and $\mathbf{B} \in \mathbb{R}^{O \times r}$ are the low-rank matrix in LoRA with $r \ll min(I, O)$. $\alpha$ represents the magnitude of the changes in $\mathbf{W}$. Only decomposed matrices $\mathbf{A}$ and $\mathbf{B}$ are updated in the fine-tuning process.

\subsection{Mixture-of-Expert (MoE-LoRA)}
The illustration of MoE-LoRA is shown in Fig.~\ref{fig:moes} (a) and (b). Each expert network in MoE-LoRA can be represented as a LoRA module worked on the FFN layer in the transformer block. MoE uses a router function to assign the learning weights of each expert in the training process of LLMs. 
% MoE can be incorporated in the PEFT of LLMs. 
% Each expert network can be represented as a LoRA module worked on the FFN layer in the transformer block. 
The coordinated weights of all experts are assigned by the router function in the feed-forward process, defined as:
\begin{equation}
\mathbf{h} = \mathbf{W}\mathbf{x} + \Delta \mathbf{W} \mathbf{x} =\mathbf{W}\mathbf{x} + \sum_{i=1}^{N}\mathcal{R}(\mathbf{x})_i\mathcal{E}_i(\mathbf{x}),
\end{equation}
where $\mathcal{E}_i$ is the $i$th expert network and $N$ is the number of experts.
% The router function $\mathcal{R}(\cdot)$ is a Softmax function that generates a probability distribution from input information that used as the weights assigned to all experts. 
The router function $\mathcal{R}(\cdot)$ employs a linear function to produce a probability distribution from input information, which serves as the weights allocated to all experts. 
Following the sparse-gated strategy in MoE studies~\citep{moe20222,li2024mixlora}, which had been proposed to address the computational overhead problem in LLMs. The top-K experts with the highest weights are selectively activated.

\section{GMoE}
The overview architecture of GMoE is shown in Fig.~\ref{fig:moes} (d). This section introduces the details of the graph router in GMoE and the propagation process in the FFN layer. Besides, to empower all experts to fully utilize their unique capabilities and keep the load balance, we use two coordination strategies: \emph{load balance strategy} and \emph{expert distinction strategy} to enhance the GMoE capability.
  
\subsection{Graph Router}
% With fewer parameters, MoE-LoRA achieves computational efficiency compared to native MoE LLMs. 
In typical MoE-LoRA literature, all experts coordinate solely depending on the simple linear router function.
% which is a simple Softmax function that assigns the weight for each expert. 
The weight assigned to each expert solely depends on the input information, with no explicit collaboration or communication mechanisms among the experts. This may result in the over-training of only a few experts and under-training of others stemming from its simplistic router strategy. The inefficient collaboration of all experts exacerbates the imbalance load problem of MoE in LLM fine-tuning. 

To enhance the collaboration of all experts, we propose a graph router to replace the simple router function. The graph router function in GMoE assigns weights to each expert in collaboration with other experts on the MoE graph. 
%functionThe weight assigned to each expert by the graph router function is computed based on the MoE graph.
Specifically, given the MoE graph $G=(V, E)$. The node set $V=\{ e_1, e_2,...e_N, x \}$ consists of all expert nodes and the input-token information node $x$ (after the self-attention layer and normalized layer in traditional transformer block). 
% As conducted in a traditional transformer, after the self-attention layer and normalized layer, the input token information 
The input node is connected to all the expert nodes to ensure all experts can receive the input information equally. The edges between all expert nodes are randomly constructed and controlled by an edge density hype parameter $\beta$.  
For the input-token node $x$ and its neighborhood expert nodes $N_e(x)$ in the MoE graph, we initialize their features using the commonly adopted Glorot uniform initialization. Subsequently, we use a graph neural network (GNN) to learn their interaction information, which not only implies the learning capability of each expert to the input-token node but also captures the collaboration information among all experts. 
% As conducted in a traditional transformer, after the self-attention layer and normalized layer,
% the input token $x$ is sent into the MoE graph $G=(V, E)$. 

The feed-forward process in GMoE can be formulated as:
\begin{equation}
\mathbf{h} = \mathbf{W}\mathbf{x} + \Delta \mathbf{W} \mathbf{x} =\mathbf{W}\mathbf{x} + \sum_{i=1}^{N}\mathcal{R}_{\textit{GNN}}(\mathbf{x})_i \mathcal{E}_i(\mathbf{x}),
\end{equation}
where $\mathbf{x}$ is the representation of input node $x$, $\mathcal{E}_i$ is the $i$th expert network, and $\mathcal{R}_{GNN}(\cdot)$ is the graph router function in MoE graph, defined as:

% The graph router function $\mathcal{R}_{GNN}(\mathbf{x})$ is defined as:
% \begin{equation}
% \mathcal{R}_{\textit{GNN}}(\mathbf{x})_i= \mathcal{R} (\mathcal{F}_{project}\textit{GNN}(e_i,N(e_i))),
% \end{equation}

\begin{equation}
\mathcal{R}_{\textit{GNN}}(\mathbf{x})_i= \mathcal{R} (\mathcal{F}(\textit{GNN}(e_i,N(e_i)))),
\end{equation}
where $\textit{GNN} (\cdot)$ is a two-layer graph neural network that learns the representation of expert node $e_i$ using information from its neighbors $N(e_i)$ that contains other experts and input token node $x$. 
$\mathcal{F}(\cdot)$ is a projection function, and
% that maps the representations from all experts into one unique vector. 
% $\mathcal{F}(\cdot)$ is the project function that maps the representations from all experts into one unique vector. 
$\mathcal{R}(\cdot)$  is the Softmax function that assigns probability weights to all experts. 

% Following the sparse-gated strategy in MoE studies~\citep{moe20222,li2024mixlora}, which had been proposed to address the computational overhead problem in LLMs.
Then we use the Top-K experts in the feed-forward process of the FFN layer, defined as:

\begin{equation}
\mathbf{h}  =\mathbf{W}\mathbf{x} + \sum_{i=1}^{K}\textit{Top-K}_{\textit{norm}} (\mathcal{R}_{\textit{GNN}}(\mathbf{x})_i \mathcal{E}_i(\mathbf{x})),
\end{equation}
where the experts with the largest top-k assigned weights from Softmax function $\mathcal{R}$ are selected, and then normalized their weights for summarizing in feed-forward operation.

The router in GMoE takes full advantage of the collaborative information aggregations by GNNs from other expert nodes. This alleviates the load unbalance and localized convergence tendencies caused by a small number of experts activated in the existing MoE studies~\citep{loadbal}. 
Except for introducing a collaboration mechanism of all experts, we use two coordination strategies, i.e., the Poisson distribution-based expert distinction strategy and the normal distribution-based load balance strategy in GMoE, to further release the capacity of each expert and increase the model stability.   

\subsection{Expert Distinction Strategy}
The key aspect of the MoE framework lies in leveraging the unique capabilities of each expert to collaborate effectively, thereby achieving enhanced overall performance.
Hence, to elicit the distinct capabilities of different experts, we optimize the assigned weights by the graph router to approximate a Poisson distribution. Specifically, for the output vector of the graph router, denoted as $\mathbf{o}_r \in \mathbb{R}^{N}$. Each dimension of $\mathbf{o}_r$ represents the assigned weight for each expert to deal with the input information. As the order of experts has no practical meaning, we sort values of each dimension in the vector $\mathbf{o}_r$ in descending order to obtain a new vector $\mathbf{v}_r \Leftrightarrow \textit{sort} (\mathbf{o}_r)$. The distribution of $\mathbf{v}_r \in \mathbb{R}^{N}$ is optimized to approximate the Poisson distribution vector $\mathbf{v}_{\textit{poisson}(\lambda,i)}$. The Kullback–Leibler (KL) divergence distance is used to calculate the loss function, defined as:

\begin{equation}
\mathbf{v}_{\textit{poisson}(\lambda,i)} = L_{norm}(\frac{\lambda^{i}}{i!}e^{-\lambda}),
\end{equation}
\begin{equation}
\textit{Loss-Poisson} =\sum_i^N \mathbf{v}_{\textit{poisson}(\lambda,i)} \log  \frac{\mathbf{v}_{\textit{poisson}(\lambda, i)}}{{\mathbf{v}_r}},
\label{poisson}
\end{equation}
where 
%$\mathbf{v}_{\textit{poisson}(\lambda,i)} = \frac{\lambda^{i}}{i!}e^{-\lambda}$, 
$i=\{1,2...,N\}$ and $N$ is the number of experts. $\mathbf{v}_{\textit{poisson}(\lambda,i)}$ is Poisson distribution vector with L1-normalization operation (i.e., $L_{norm}$) and $\lambda$ is the learning parameter.

\subsection{Load Balance Strategy}
Apart from the distinct capabilities of different experts, one important factor in MoE training is keeping the load balance of all experts. Otherwise, greater routing weights on a small number of experts in the early stages of the fine-tuning process will result in a rapid localized optimization problem. In GMoE, we propose the Normal distribution-based load balance strategy.
Specifically, as only Top-K experts are activated in dealing with input information, we calculate the cumulative weights of each expert to represent its activation frequency. 
Then construct the activation frequency vector of all experts by normalizing the cumulative weights, denoted as $\mathbf{v}_a$ in the next feed-forward step. 
Our aim is to make the activation frequency vector $\mathbf{v}_a$ follow a natural normal distribution rather than the absolute equality in existing MoE literature~\citep{li2024mixlora,moe20222}. 
Formally, the Normal distribution-based load balance loss function can be defined as:

% Different from existing studies in which an auxiliary loss proposed by Switch Transformers~\citep{switch-trans} to keep the load of each experts balance, an advanced normal distribution loss is propsed in our paper, which relaxing the restrictions of multiple experts and keep the utilization more closer to natural patterns.

\begin{equation}
\mathbf{v}_{\textit{normal}(\mu,\sigma,i)} = L_{norm} (\frac{1}{\sqrt{2\pi\sigma}} \exp(-\frac{{(i-u)}^2}{2\sigma^2})),
\end{equation}
\begin{equation}
\textit{Loss-Normal} =\sum_i^N \mathbf{v}_{\textit{normal}(\mu,\sigma,i)} \log  \frac{\mathbf{v}_{\textit{normal}(\mu, \sigma,i)}}{{\mathbf{v}_a}},
\label{norm_loss}
 \end{equation}
where
%$\mathbf{v}_{\textit{normal}(\mu,\sigma,i)} = \frac{1}{\sqrt{2\pi}} \exp(-\frac{{i-u}^2}{2\sigma^2})$, 
$i=\{1,2...,N\}$ and $N$ is the number of experts. $\mathbf{v}_{\textit{normal}(\mu,\sigma,i)}$ is the normalized vector that follows the Normal distribution, $\mu =\frac{N}{2}$ is the mean of all samples, and $\sigma$ is learning parameter optimized in the fine-tuning process.

\section{Experiments}

\subsection{Experimental Settings}

\paragraph{Datasets.}
To evaluate the performance of our framework, we use four representative public datasets, including the question-answering task, i.e., ARC-Challenge~\citep{arc}, OpenBookQA~\citep{obqa}, SIQA~\citep{siqa}, and task classification task in BoolQ~\citep{boolq} dataset. These datasets cover the evaluations on different domains of LLMs, such as factual knowledge from Wikipedia, natural science, science facts, and social interactions.
% All tasks in four datasets are evaluated using the accuracy metric.  
% The statistics of the datasets are shown in Table~\ref{tab:dataset}.

% \begin{itemize}
%     \item{\textbf{ARC-Challenge}}. It is
%     released by the Allen Institute for AI and is widely used to evaluate the reading comprehension and reasoning abilities of LLMs. The dataset consists of multiple-choice questions derived from U.S. elementary school science exams. 
%     \item{\textbf{BoolQ}}. It is a question-answering dataset and contains Yes/No type answers extracted from real search queries. Each question is described by a relevant passage of text.
%     This dataset evaluates the natural language understanding capability of LLMs, particularly in handling short-form question answering and textual reasoning.
%     \item{\textbf{OpenBookQA}}. It is designed to evaluate the ability of LLMs to understand elementary-level science knowledge. It consists of multiple-choice questions. Each question is accompanied by a small set of "facts" that are required to answer the questions. 
%     \item{\textbf{SIQA}}. Social Interaction QA is a benchmark for multiple-choice question answering focused on social intelligence reasoning. It consists of questions that require models to understand and reason about social situations, such as interpreting human behavior, emotions, and intentions. Each question contains a brief introduction with one correct answer in three answer choices.
% \end{itemize}

\begin{table*}
    \centering
    \small
    \caption{The comparisons of model \textbf{Accuracy} and \textbf{Stability} measured by standard deviation (Std). The smaller the standard deviation, the greater the stability of the model. The underline represents the SOTA baseline method. The best results on Accuracy and Stability are highlighted in bold. 
    For Stability evaluation, the superior outcomes of our approach compared with sparse MoE methods (excluding LoRAMoE) are emphasized in italics.
    %the best results in sparse MoE methods (i.e., except LoRAMoE) are highlighted in italics. 
    % the dense MoE model LoRAMoE is relatively stable due to all experts being activated. So we compare our method with sparse MoE methods (i.e., except LoRAMoE), and highlighted the best results in italics.
    }
    \begin{tabular}{ll|ccccc|ccccc}
    \toprule
        &&\multicolumn{5}{|c|}{\textbf{Accuracy Evalution ($\uparrow $)}}&\multicolumn{4}{c}{\textbf{Stability Evaluation (Std$\downarrow$)}}\\
       \cline{3-12}
         LLMs& Method & ARC-C & BoolQ & OBQA & SIQA & Avg. & ARC-C & BoolQ & OBQA & SIQA & Avg.\\
         \midrule
         \multirow{6}{*}{Llama3} %& LoRA & 77.31 & 74.95 & 86.93 & \underline{79.73} & 1.49 & 1.01 & 1.7 & 0.39\\
         & LoRAMoE & 76.77 & 74.48& \underline{87.33} & \underline{79.66} & 79.56 & 0.25 & 0.66 & \underline{1.01} & 0.39 & 0.58\\
         & MING-MoE & 77.28& 72.95 & 86.47& 79.48 & 79.05 & 0.91 & 0.91 & 1.42 & 0.31 & 0.89\\
         & MoLA & 76.74 & 73.50 & 84.00 & 78.70 & 78.24 & \underline{0.47} & \underline{0.34} & 1.44 & 0.90 & 0.79\\
         & MixLoRA & \underline{77.36}& \underline{75.43} & 87.20 & 79.53 & \underline{79.88} & 0.52 & 0.99 & {1.13} & \textbf{0.15} & \underline{0.70}\\
         & \textbf{GMoE} & \textbf{77.56}& \textbf{75.90} &  \textbf {88.13} & \textbf{80.48} & \textbf{80.52} & \textbf{\emph{0.26}} & \textbf{0.17} & \textbf{0.98} & 0.37 & \textbf{0.45}\\
         \midrule
         \multirow{6}{*}{Qwen2}% & LoRA & 82.68& 74.03 & \underline{91.2} & 80.44 & 0.82 & \underline{0.47} & \underline{0.35} & 2.21\\
         & LoRAMoE & 83.50& \underline{74.80} &90.53& 80.54 & 82.34 & 0.44 & 0.03 & 0.23 & 1.15 & 0.46\\
         & MING-MoE & 83.99 & 74.21 &\underline{91.00} & \underline{80.67} & \underline{82.47} & 1.51 & \underline{0.62} & 2.00 & 1.12 & 1.31\\
         & MoLA & 83.19 & 74.48 & 89.93 & 80.59 & 82.05 & 1.34 & 0.75 & 1.79 & \underline{0.83} & 1.18\\
         & MixLoRA & \underline{84.41} & 74.77 & 90.00 & 80.31 & 82.37 & \underline{0.69} & 0.90 & \underline{0.69} & 1.90 & \underline{1.04}\\
         & \textbf {GMoE} &  \textbf{85.10} & \textbf{75.40} & \textbf {91.67} & \textbf {80.98} & \textbf{83.29} & \textbf{\emph{0.48}} & \textbf {\emph{0.57}} & \textbf {\emph{0.64}} & \textbf{0.24} & \textbf{0.48}\\
         \midrule
         \multirow{6}{*}{Yi-1.5}% & LoRA & \underline{84.76}& 72.99 & 90.33 & 81.65 & \textbf {0.24} & 2.19 & 1.62 & 1.22\\
         & LoRAMoE & 84.33& 72.89 & 91.73 & 80.94 & 82.47 & 3.09 & 0.38 & 0.42 & 0.36 & 1.06\\
         & MING-MoE & \underline{84.58} & 73.32 & 90.07& 81.56 & 82.38 & 0.61 & 0.58 & 0.50 & 0.39 & \underline{0.52}\\
         & MoLA & 84.47 & 72.26 & 90.07 & 81.25 & 82.01 & \underline{0.61} & \textbf{0.21} & 1.03 & 0.31 & 0.54\\
         & MixLoRA &  84.36 & \underline{73.32} & \textbf{91.80} & \underline{81.89} & \underline{82.84} & 1.75 & 0.39 & \underline{0.40} & \underline{0.31} & 0.71\\
         & \textbf {GMoE} & \textbf {85.32} & \textbf {74.23} & 91.33& \textbf{82.24} & \textbf{83.28} & \textbf{0.52} & 0.32 & \textbf{0.40} & \textbf{0.28} & \textbf{0.38}\\
         \bottomrule
    \end{tabular}
    \label{tab:result}
\end{table*}

\paragraph{Baseline Methods.}

To verify the effectiveness of GMoE, we compare the performance of typical PEFT MoE methods on three popular open-source LLMs, i.e., Llama3-8B\footnote{https://github.com/meta-llama/llama3}, Qwen2-7B~\citep{qwen2}, and Yi-1.5-9B~\citep{yi}.
% The baseline methods are introduced as follows.

%\item{\textbf{LoRA}}~\citep{lora}. It is the most popular parameter-efficient technique in LLM fine-tuning. It adopts a low-rank decomposition technique to update the parameters in the decomposed matrix.
\paragraph{LoRAMoE~\citep{loramoe}.} It is a representative dense MoE model. A localized balancing constraint is used to alleviate the knowledge-forgetting problem in the model updating process.
\paragraph{MING-MoE~\citep{liao2024ming}.} It is a typical sparse MoE architecture without constant loss functions and designed for medical multi-task learning. 
\paragraph{MoLA~\citep{mola}.} It is a recently proposed MoE-based PEFT model that applies different numbers of experts in router functions in different layers. More experts are used at higher layers of the transformer block. 
% We reproduce this model under mlora architecture and make sure the total number of experts in MoLA is the same as the others.
\paragraph{MixLoRA~\citep{li2024mixlora}.} It is the SOTA method of PEFT in MoE literature. 
It adopts the MoE-LoRA architecture in the FFN layer of LLMs. To address the imbalance load problem, it uses an average auxiliary load balance loss.
\paragraph{GMoE.} Different from MixLoRA, we propose a novel graph router. It takes advantage of the collaboration of all experts learned by graph neural networks. Besides, two coordination strategies, i.e., the expert distinct strategy and load balance strategy, are proposed to enhance the capabilities of all experts.

% \begin{itemize}
% %\item{\textbf{LoRA}}~\citep{lora}. It is the most popular parameter-efficient technique in LLM fine-tuning. It adopts a low-rank decomposition technique to update the parameters in the decomposed matrix.
% \item{\textbf{LoRAMoE}}~\citep{loramoe}. It is a representative dense MoE model. A localized balancing constraint is used to alleviate the knowledge-forgetting problem in the model updating process.
% \item{\textbf{MING-MoE}}~\citep{liao2024ming}. It is a typical sparse MoE architecture without any constant loss function and is designed for medical multi-task learning. 
% \item{\textbf{MoLA}}~\citep{mola}. It is a recently proposed MoE-based PEFT model that applies different numbers of experts in router functions in different layers. More experts are used at higher layers of the transformer block. 
% % We reproduce this model under mlora architecture and make sure the total number of experts in MoLA is the same as the others.
% \item{\textbf{MixLoRA}}~\citep{li2024mixlora}. It is the SOTA method of PEFT in MoE literature. 
% It adopts the MoE-LoRA architecture in the FFN layer of LLMs. To address the imbalance load problem, it uses an average auxiliary load balance loss.
% \item{\textbf{GMoE}}. Different from MixLoRA, we propose a novel graph router. It takes advantage of the collaboration of all experts learned by graph neural networks. Besides, two coordination strategies, i.e., the expert distinct strategy and load balance strategy, are proposed to enhance the capabilities of all experts.
% \end{itemize}

\paragraph{Parameter Settings.}
For each baseline method, a grid search is applied to find the optimal settings.
These include learning rate from $\{0.1, 0.01, 0.001, 0.0001, 0.00001\}$, number of experts from $\{4, 8, 12, 16\}$, rank of LoRA from $\{1, 2, 4, 8, 16\}$, Top-K experts from $\{1, 2, 3, 4, 5\}$.
We report the results of each method with its optimal hyperparameter settings on the validation data. 
In our model, we adopt GCN as the graph neural network (GNN). The number of aggregation layers in GCN is 2, and the hidden dimension of GCN layers is 256. The edge density $\beta$ to construct the MoE graph is $0.1$. The $\alpha$ in Eq.~\ref{equ:lora} is 4 in the LoRA component.  The coefficients of the Poisson distribution loss function and Normal distribution loss function are set to 0.005 and 8 in the final loss function in LLM fine-tuning. 
% The implementation code will be available after the review process.  

% Throughput based on Llama3-8B with an A800-80G.
% More details of the experimental implementation is available at~\url{https://github.com/BAI-LAB/GMoE}.

\begin{figure*}
\centering
\setcounter{subfigure}{0}
\begin{subfigure}[t]{0.24\linewidth}  % 调整宽度为约1/4页面宽度
    \centering
    \includegraphics[width=\linewidth]{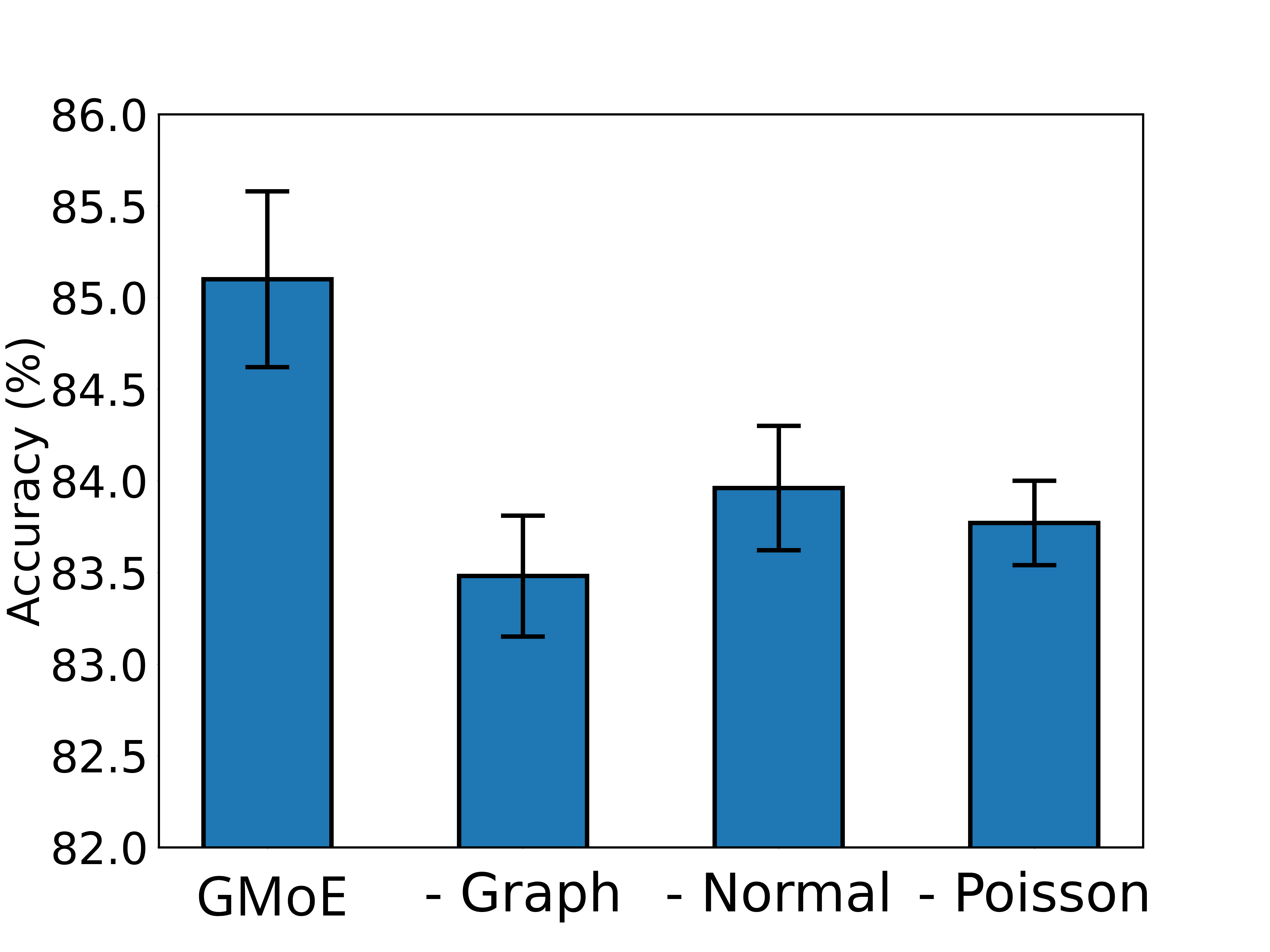}
    \caption{ARC-Challenge}
    \label{fig:abli1}
\end{subfigure}
\hfill
\begin{subfigure}[t]{0.24\linewidth}
    \centering
    \includegraphics[width=\linewidth]{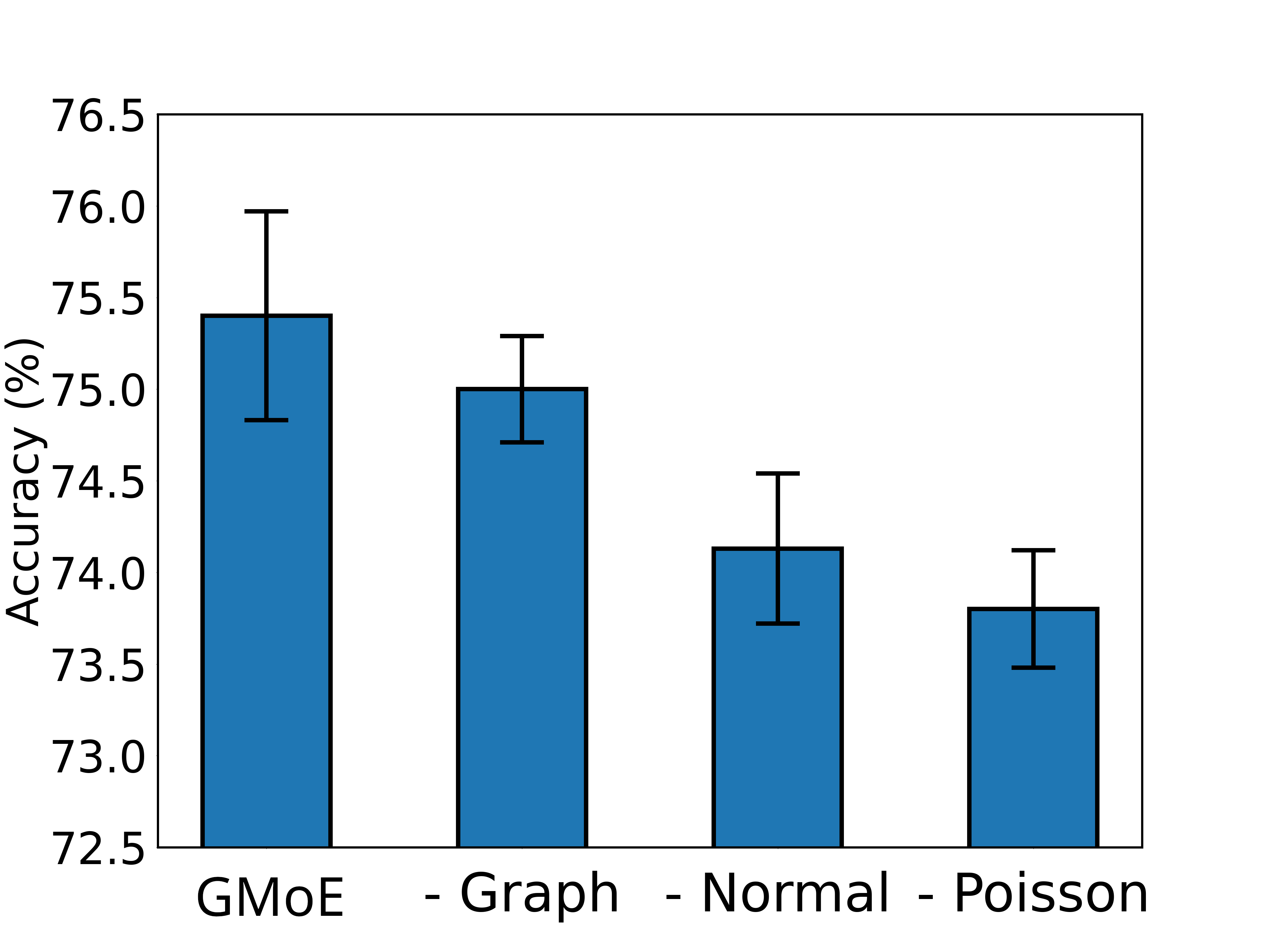}
    \caption{BoolQ}
    \label{fig:abli2}
\end{subfigure}
\hfill
\begin{subfigure}[t]{0.24\linewidth}
    \centering
    \includegraphics[width=\linewidth]{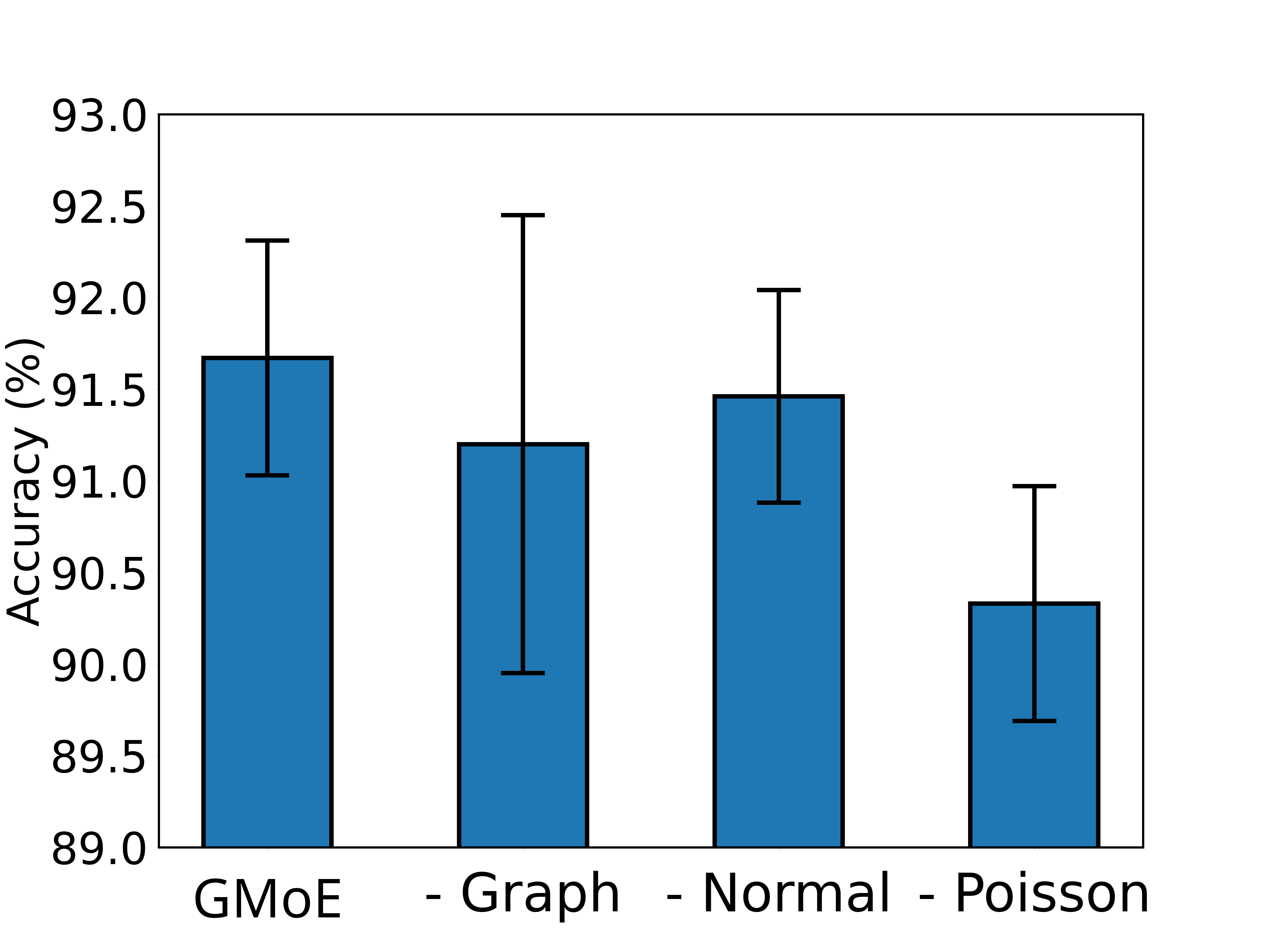}
    \caption{OpenBookQA}
    \label{fig:abli3}
\end{subfigure}
\hfill
\begin{subfigure}[t]{0.24\linewidth}
    \centering
    \includegraphics[width=\linewidth]{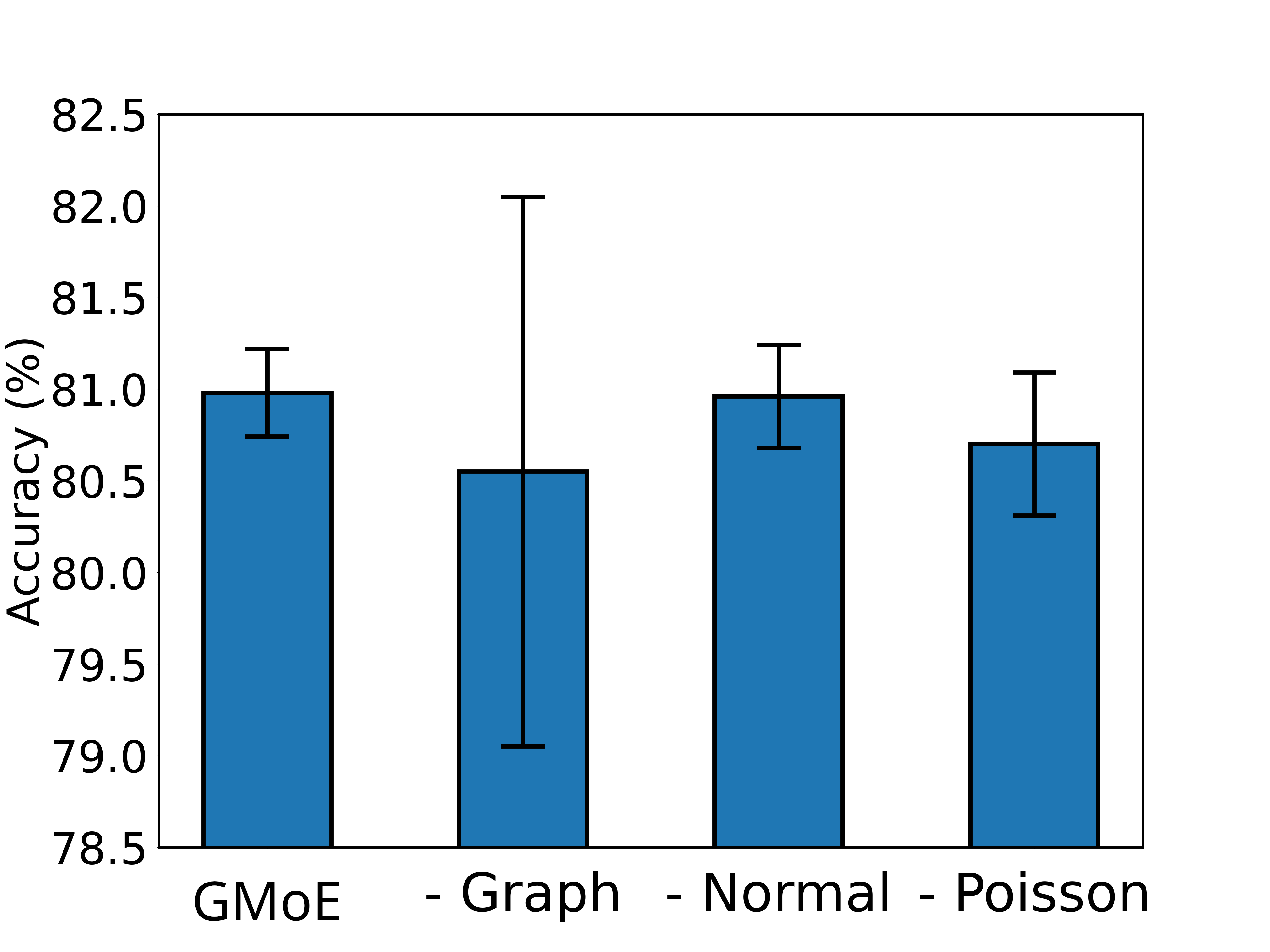}
    \caption{SIQA}
    \label{fig:abli4}
\end{subfigure}
\caption{The model performance of degradation variants of GMoE on Qwen2-7B.}
\label{fig:abli}
\end{figure*}

\begin{figure*}
    \centering
    \setcounter{subfigure}{0}
    % 第一张子图
    \begin{subfigure}[h]{0.19\linewidth}
        \centering
        \includegraphics[width=\linewidth]{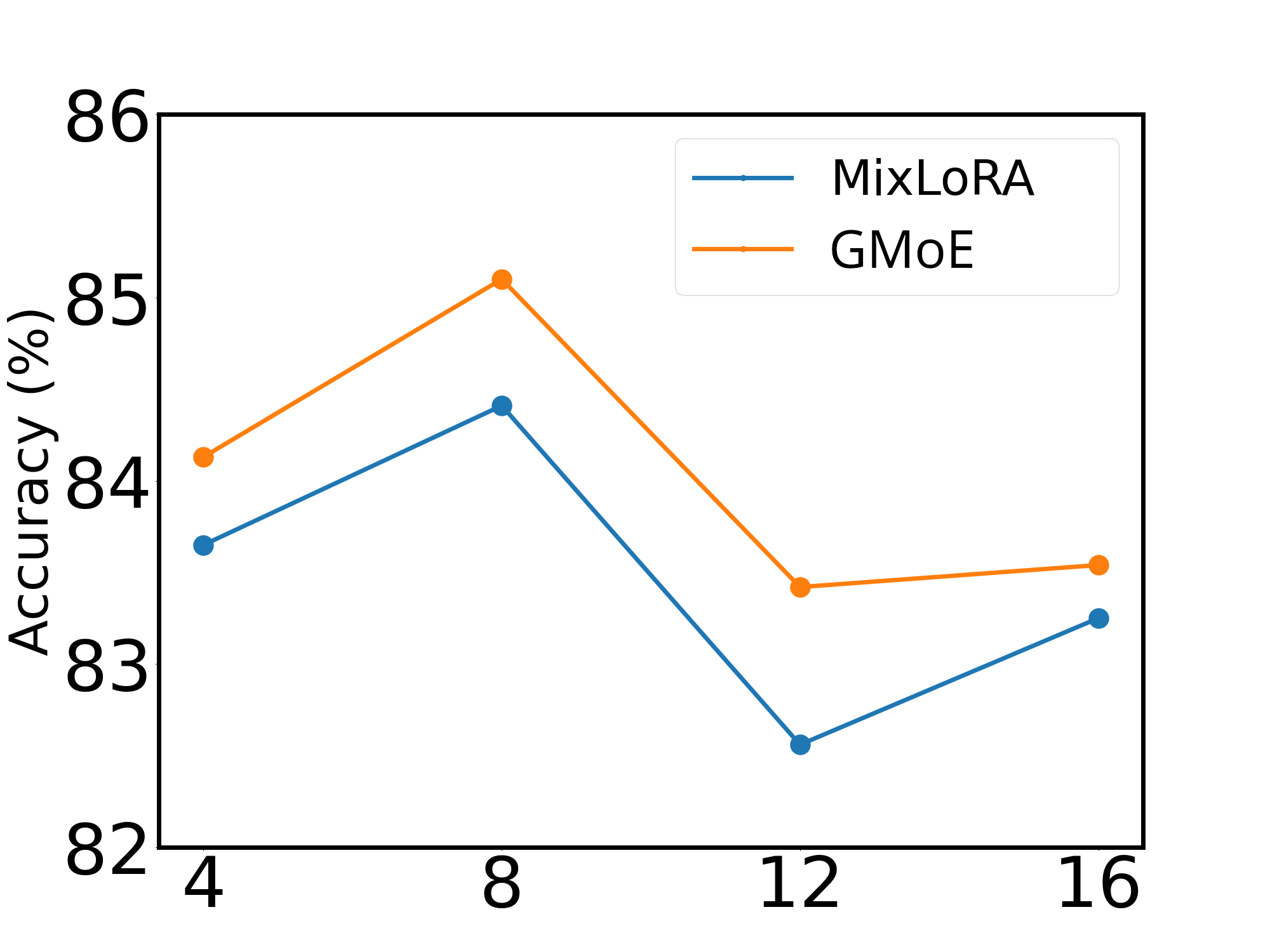}
        \caption*{Expert number}
        \label{fig:exp1}
    \end{subfigure}
    \hfill
    % 第二张子图
    \begin{subfigure}[h]{0.19\linewidth}
        \centering
        \includegraphics[width=\linewidth]{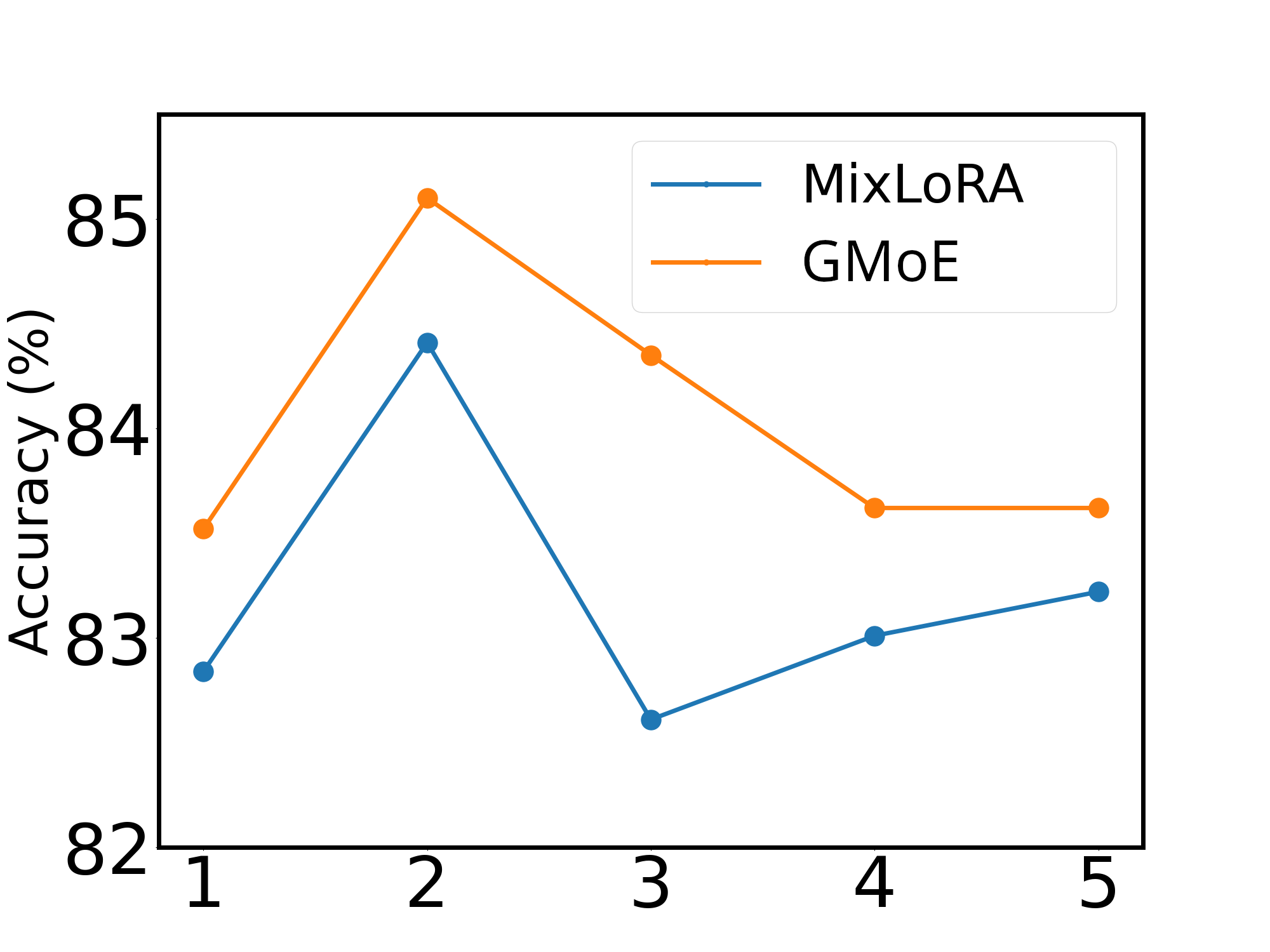}
        \caption*{Top-K}
        \label{fig:topk1}
    \end{subfigure}
    \hfill
    % 第三张子图
    \begin{subfigure}[h]{0.19\linewidth}
        \centering
        \includegraphics[width=\linewidth]{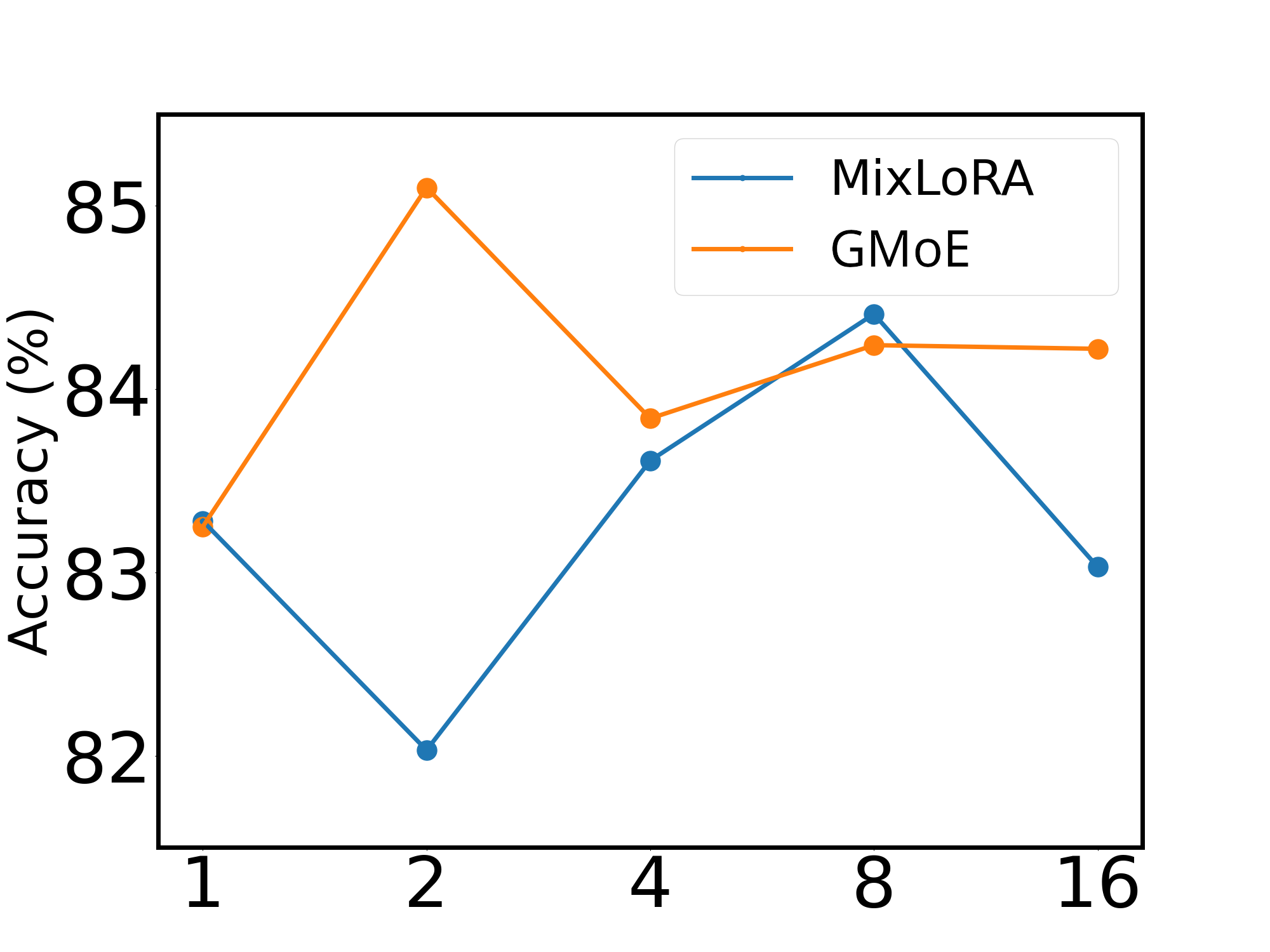}
        \caption*{Rank of LoRA}
        \label{fig:rank1}
    \end{subfigure}
    \hfill
    \begin{subfigure}[h]{0.19\linewidth}
        \centering
        \includegraphics[width=\linewidth]{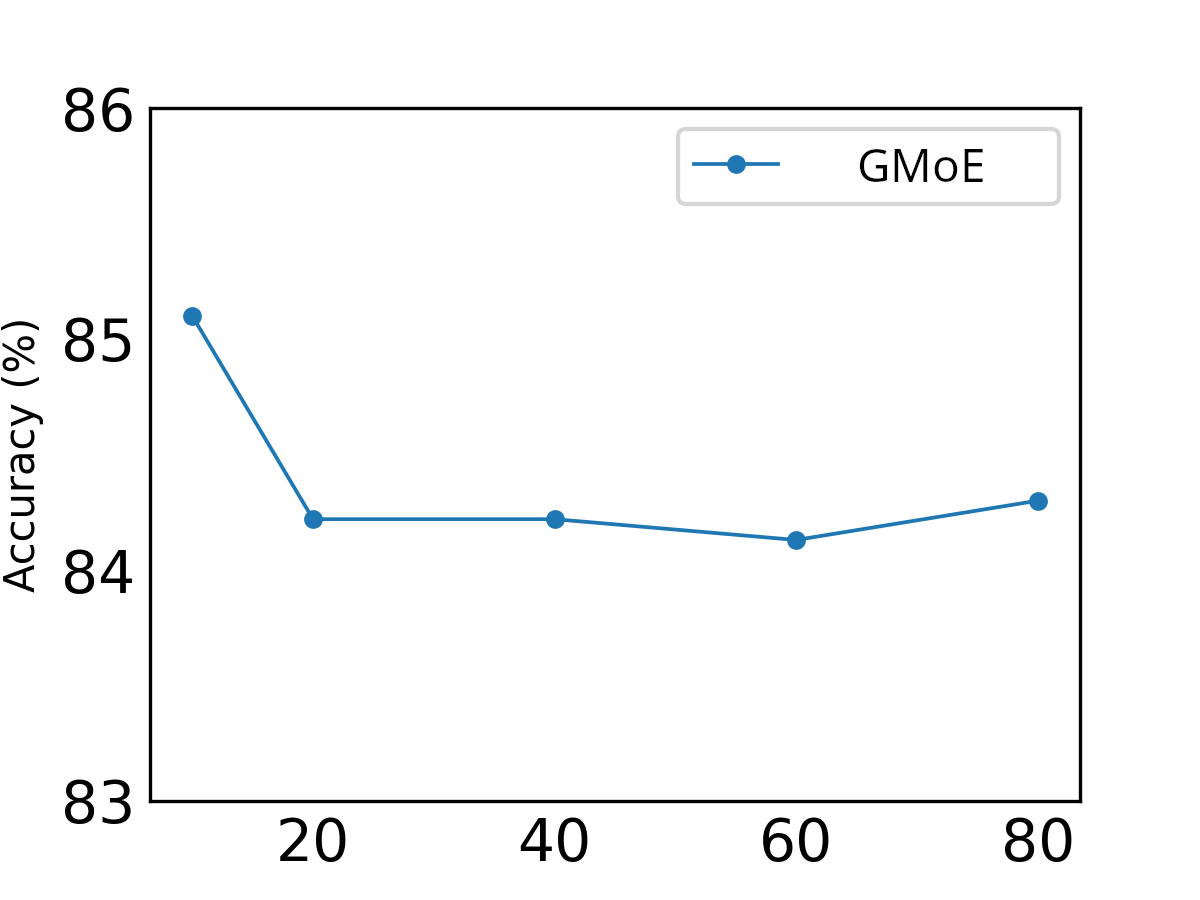}
        \caption*{Edge density}
        \label{fig:edge1}
    \end{subfigure}
    \hfill
    \begin{subfigure}[h]{0.19\linewidth}
        \centering
        \includegraphics[width=\linewidth]{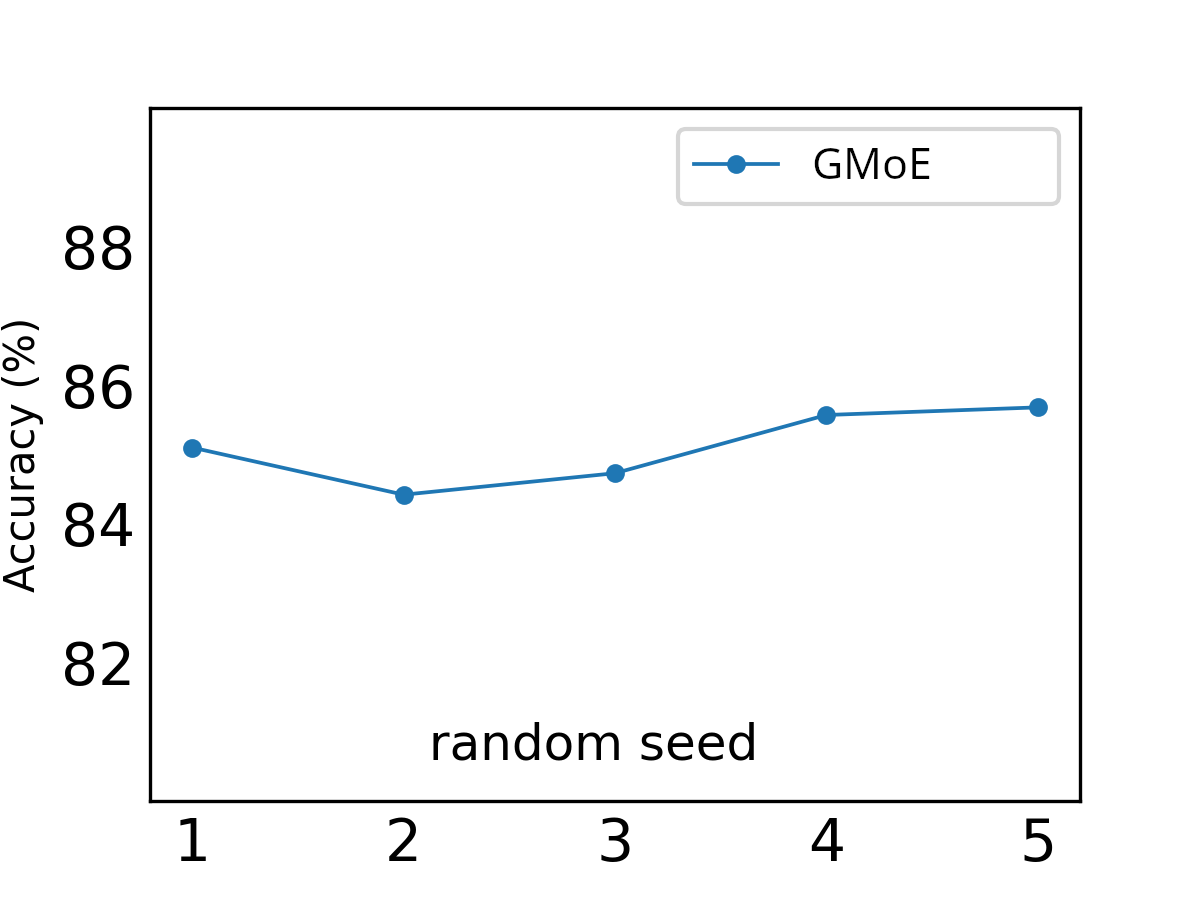}
        \caption*{Initial edge}
        \label{fig:nonsense}
    \end{subfigure}

   % \vspace{0.3cm}
    {(a) ARC-Challenge Dataset} % 小标题
  %   \vspace{0.5cm}

    % 第一张子图
    \begin{subfigure}[h]{0.19\linewidth}
        \centering
        \includegraphics[width=\linewidth]{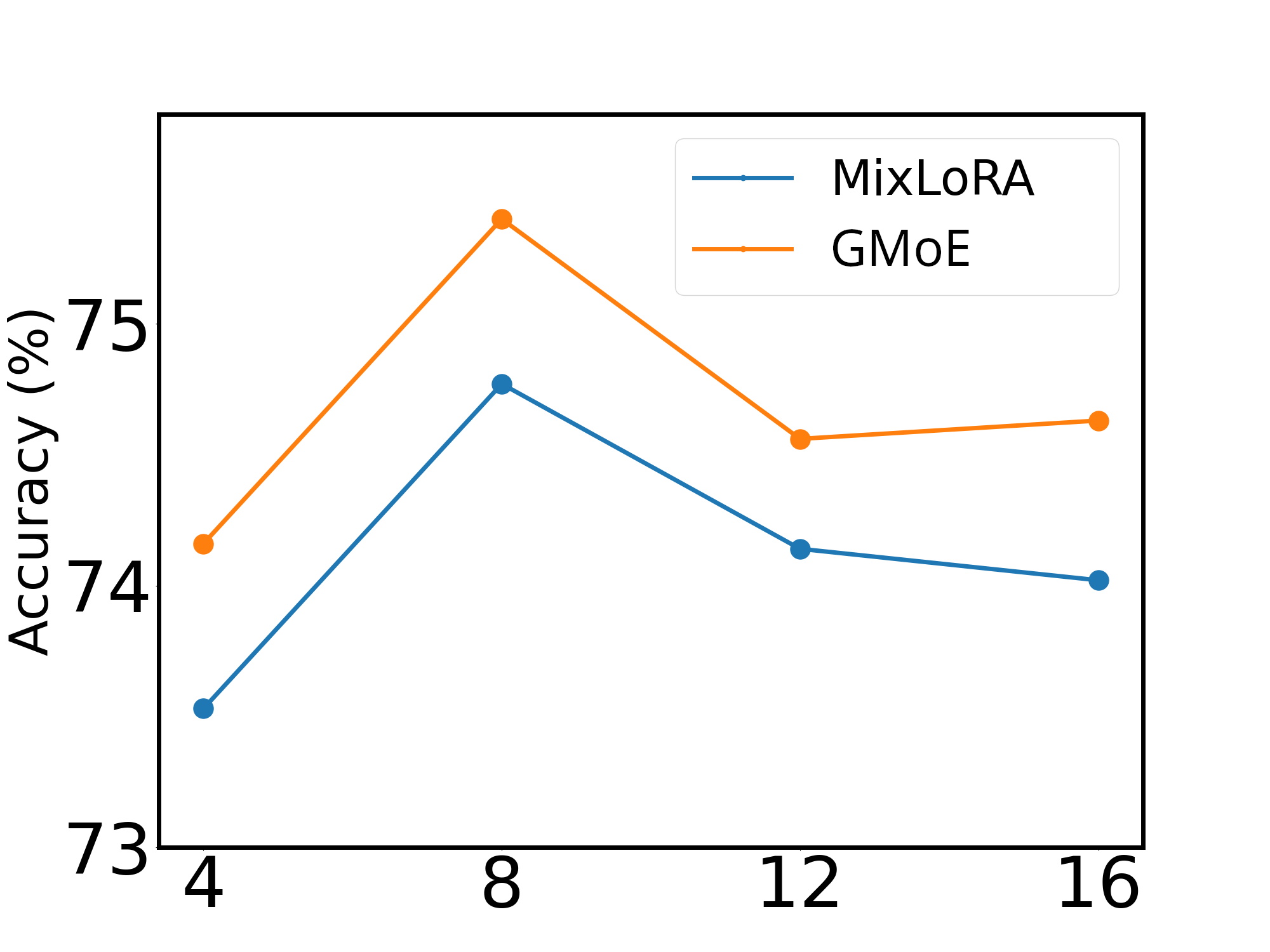}
        \caption*{Expert number}
        \label{fig:exp2}
    \end{subfigure}
    \hfill
    % 第二张子图
    \begin{subfigure}[h]{0.19\linewidth}
        \centering
        \includegraphics[width=\linewidth]{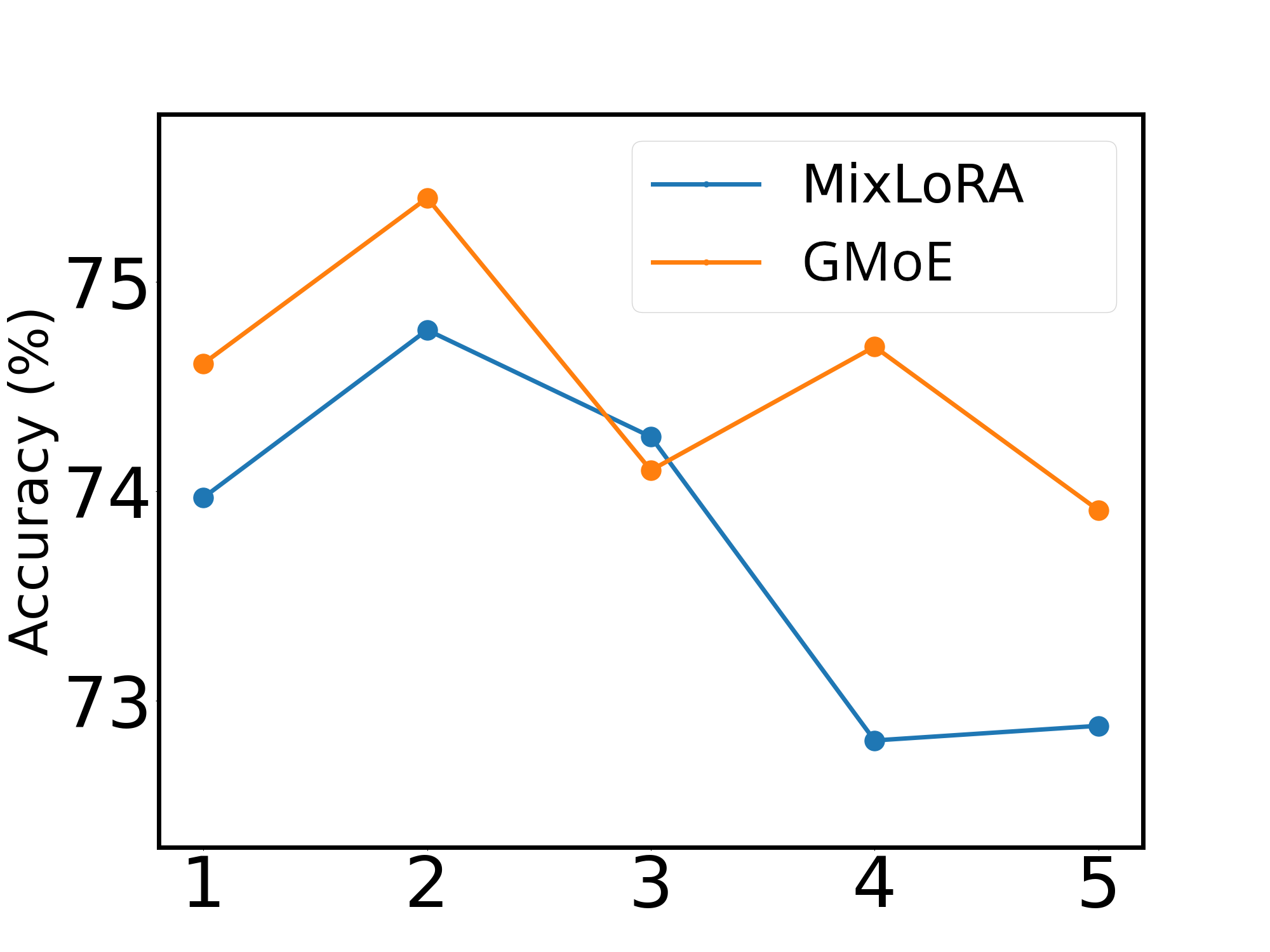}
        \caption*{Top-K}
        \label{fig:topk2}
    \end{subfigure}
    \hfill
    % 第三张子图
    \begin{subfigure}[h]{0.19\linewidth}
        \centering
        \includegraphics[width=\linewidth]{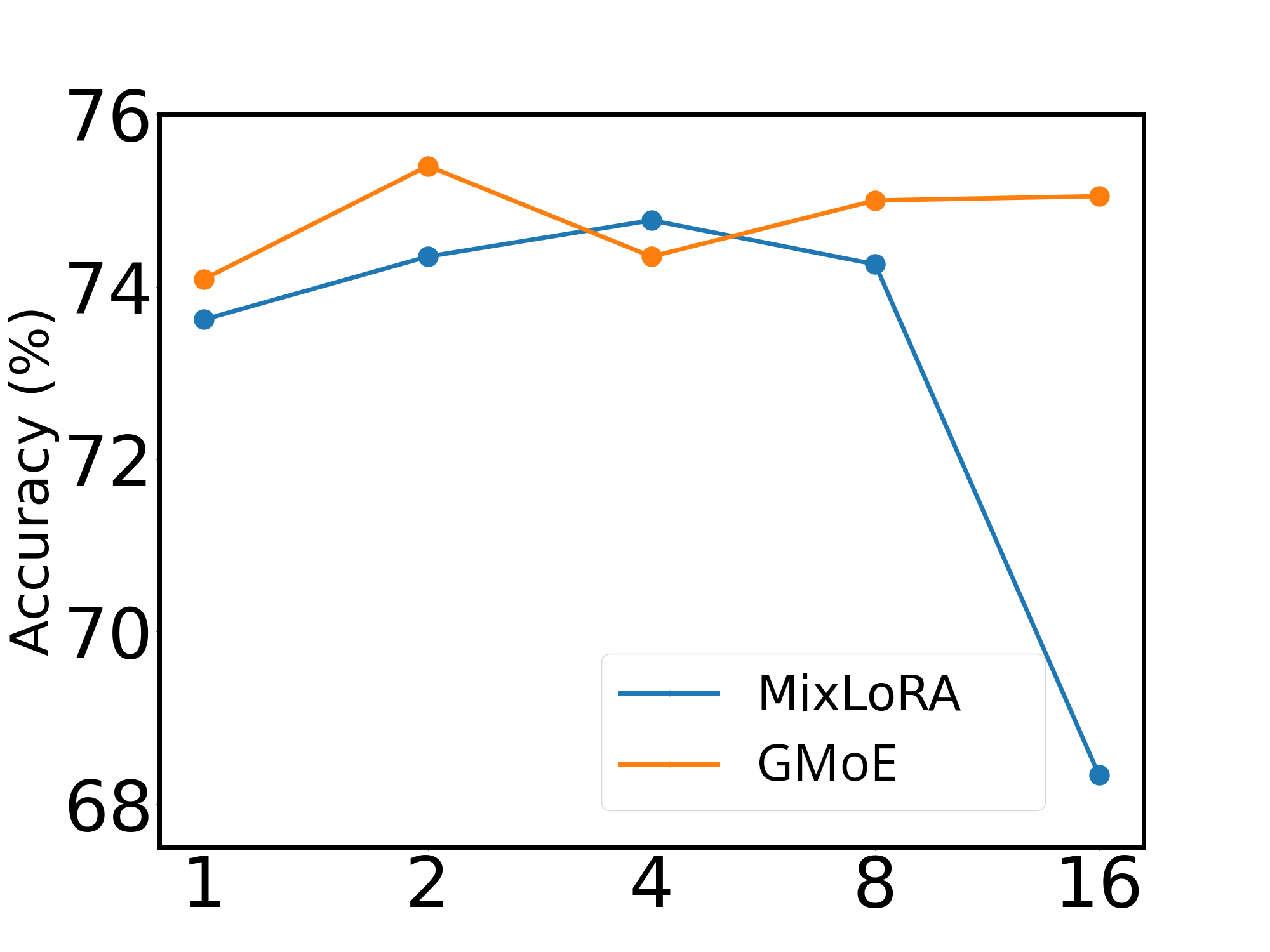}
        \caption*{Rank of LoRA}
        \label{fig:rank2}
    \end{subfigure}
    \hfill
    \begin{subfigure}[h]{0.19\linewidth}
        \centering
        \includegraphics[width=\linewidth]{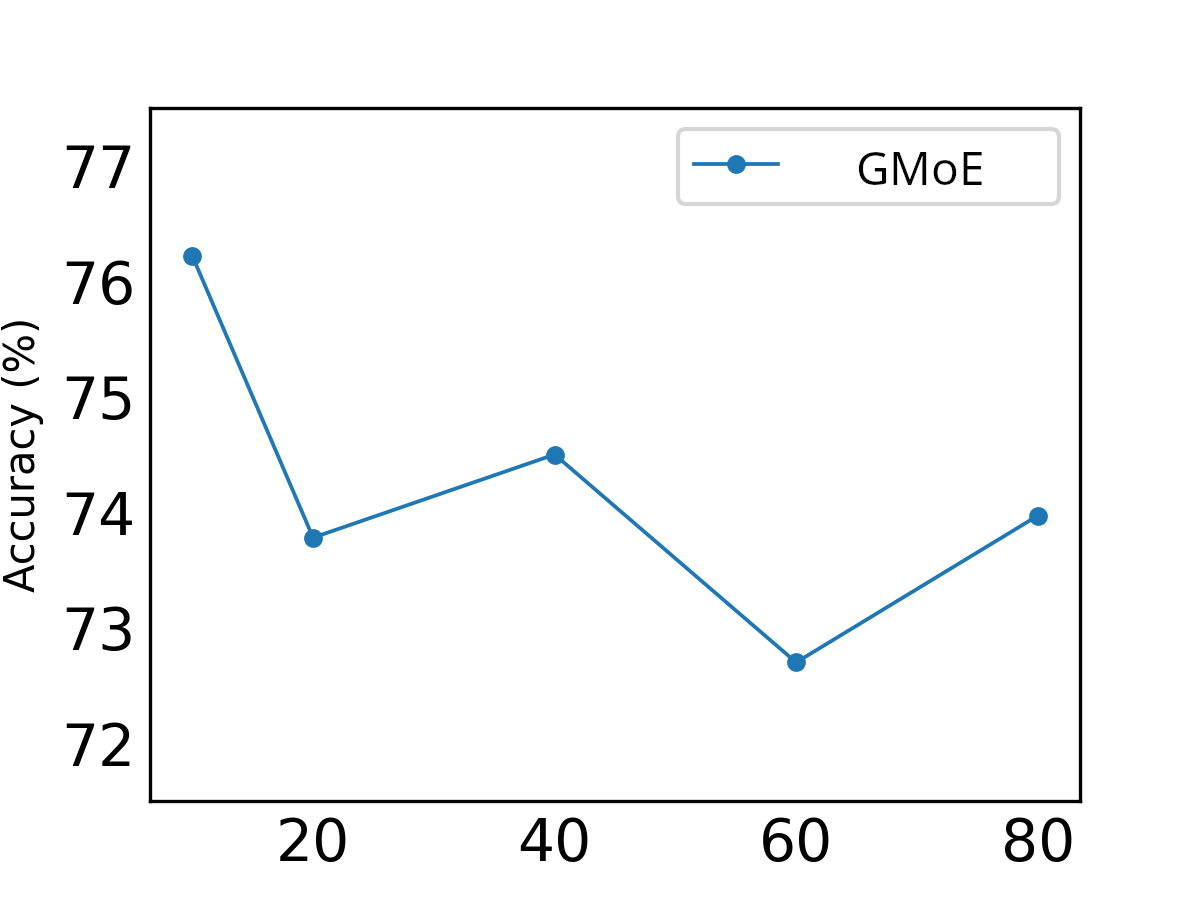}
        \caption*{Edge density (\%)}
        \label{fig:edge2}
    \end{subfigure}
    \hfill
    \begin{subfigure}[h]{0.19\linewidth}
        \centering
        \includegraphics[width=\linewidth]{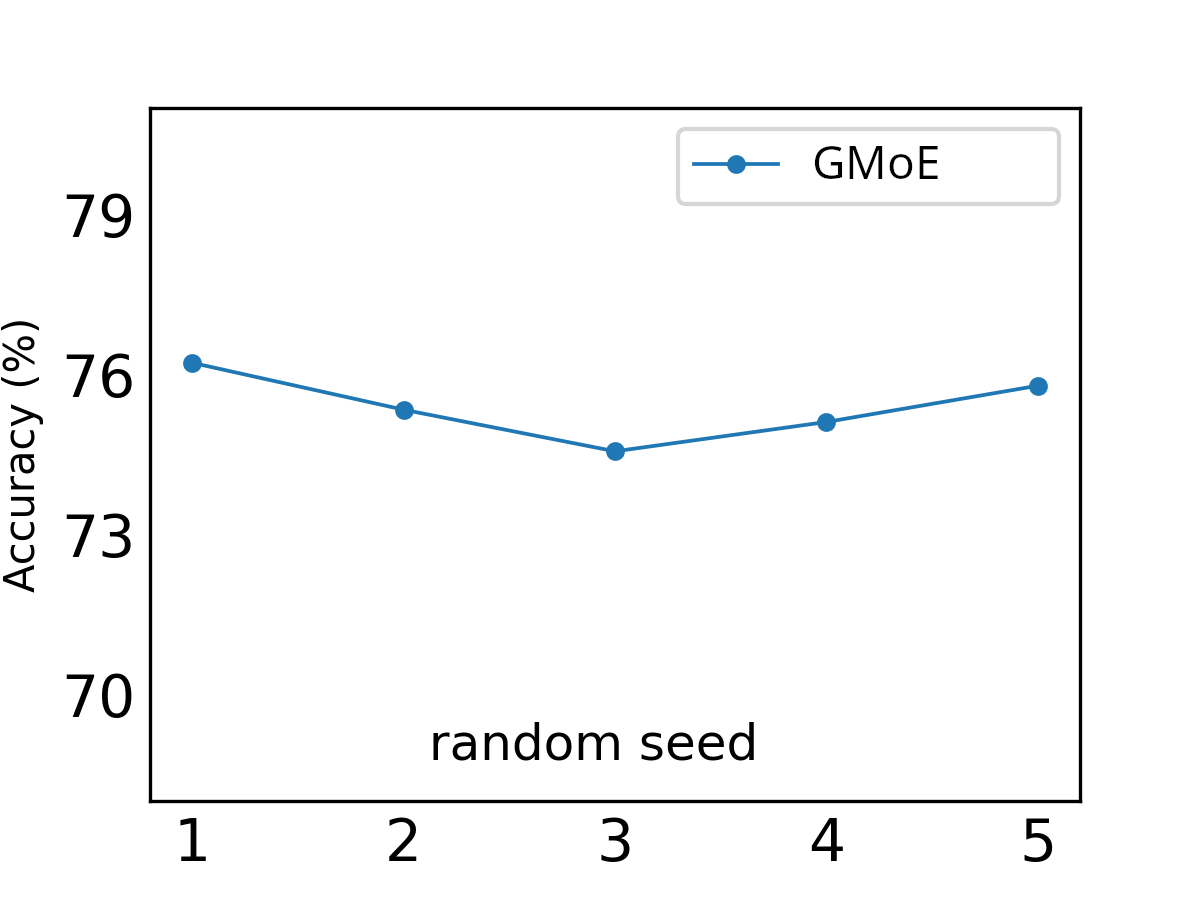}
        \caption*{Initial edge}
        \label{fig:nonsense2}
    \end{subfigure}
   % \vspace{0.3cm}
    (b) BoolQ Dataset % 小标题
   % \vspace{0.5cm}    
    \caption{The hyper-parameters analysis in ARC-Challenge and BoolQ dataset based on Qwen2-7B.}
    \label{fig:para}
\end{figure*}

\subsection{Main Results}

% We make comprehensive evaluations of different MoE methods from three aspects, i.e.,  \textbf{Accuracy}, \textbf{Stability}, and \textbf{Efficiency}.
We make comprehensive evaluations of different MoE methods from both \textbf{Accuracy} and  \textbf{Stability} evaluations.
%By enhancing the MoE collaborations in graph router, our GMoE release the capability of each expert and meanwhile alleviate the imbalance load problem.
% To make comprehensive evaluations, we compare the performance of different MoE method from three aspects, i.e.,  \textbf{Accuracy}, \textbf{Stability} and \textbf{Efficiency}.
We repeat each experiment under five different random seeds, and report the mean value of accuracy. The standard deviation of each model is used to measure the stability of different models. 
% The results on four datasets with three different base LLMs are shown in Table~\ref{tab:result} and Fig.~\ref{fig:scatter}. 
% The efficiency is evaluated by the size of learning parameters in the fine-tuning process of LLMs.
% We have the following observations.
% \subsubsection{Accuracy \& Stability  Evaluation} 
% We repeat experiments under five different random seeds, and report the mean value of the model accuracy.
%The standard deviation of each model is used to measure the stability of method. The efficiency is evaluated by the size of learning parameters in the fine-tuning process of LLMs.
% The biggest challenge in MoE fine-tuning research is the instability of model performance caused by the imbalance load problem.
% The accuracy and stability evaluations are 
As shown in Table~\ref{tab:result}. The smaller the standard deviation, the greater the stability of the model. We can see that:
% Model performance on accuracy metric is shown in Table~\ref{tab:result}. We can see that:

(1)  The sparse MoE architecture with an auxiliary load balance loss in MixLoRA performs better than both the sparse MoE (i.e., MING-MoE, MoLA) and dense MoE methods (i.e., LoRAMoE) in many cases, showing the importance of enhancing the coordination of experts.

(2) The performance of the methods varies across different datasets. For example, for the easier task with higher model accuracy in the OpenBookQA dataset, the dense MoE method (i.e., LoRAMoE) performs better in Llama3 and Yi-1.5.
% and the simple LoRA method shows advantages over other baselines in Qwen2.
This indicates that localized optimization of sparse MoE may reduce the model's capability and lose its advantages in easy task learning. 

(3) GMoE achieves the best performance in most cases on four datasets, showing the effectiveness of the collaborations of multiple experts learned on the MoE graph.
% With expert collaboration in GMoE, our method consistently exhibits the best performance in various downstream tasks.

(4) Among sparse MoE methods (MING-MoE, MoLA, MixLoRA), GMoE exhibits the smallest standard deviation in most cases, achieving stability comparable to LoRAMoE, which is a dense MoE activating all experts for balance but with high computational costs.
% This demonstrates GMoE’s superiority in addressing load imbalance efficiently, bypassing the trade-offs of dense architectures.

\begin{figure*}[htbp]
\centering
\small
\setcounter{subfigure}{0}
\begin{subfigure}[h]{0.24\linewidth}
    \centering
    \includegraphics[width=\linewidth]{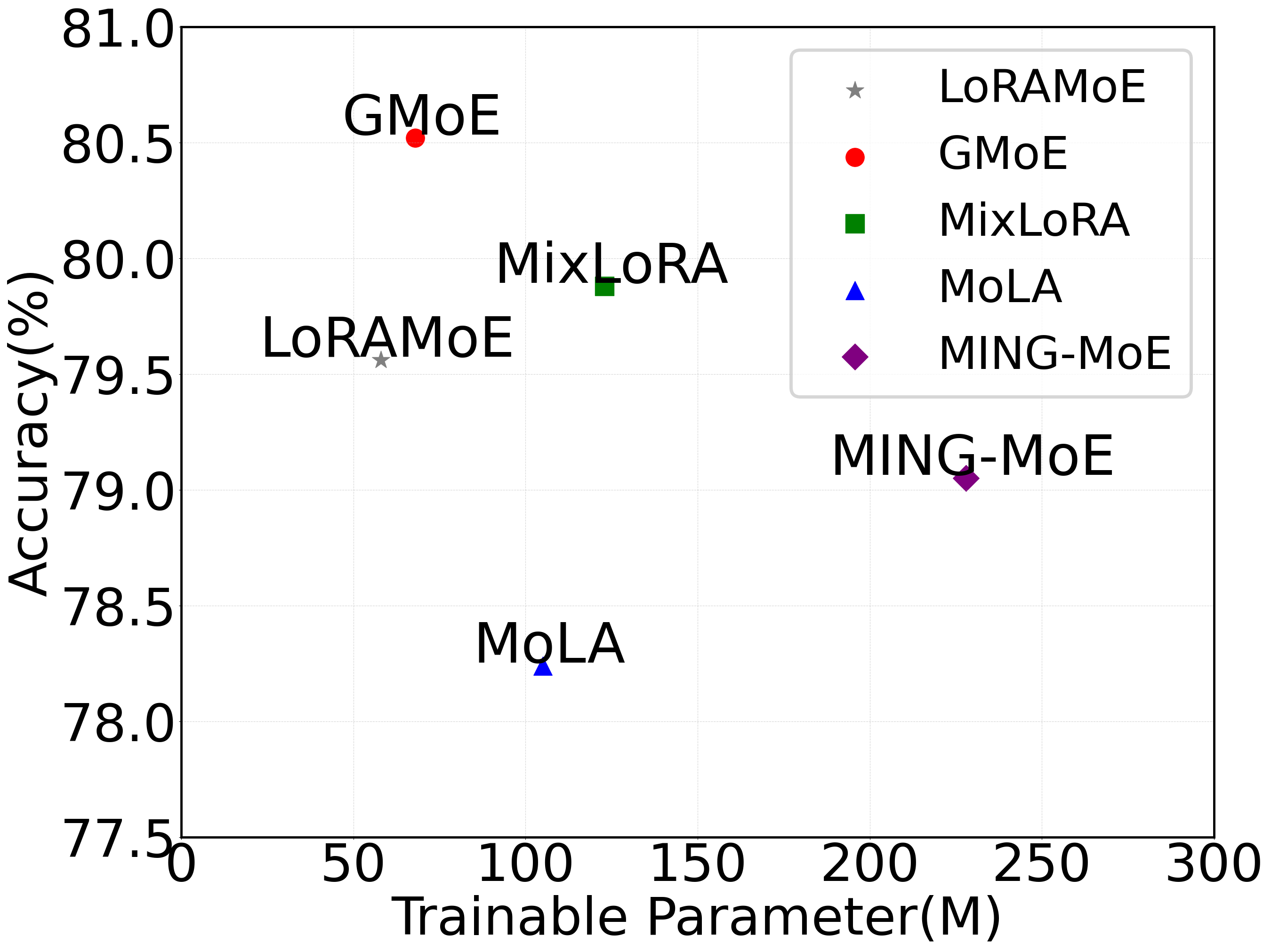}
    \caption{Llama3-8B}
    \label{fig:sca1}
\end{subfigure}
\hfill
\begin{subfigure}[h]{0.24\linewidth}
    \centering
    \includegraphics[width=\linewidth]{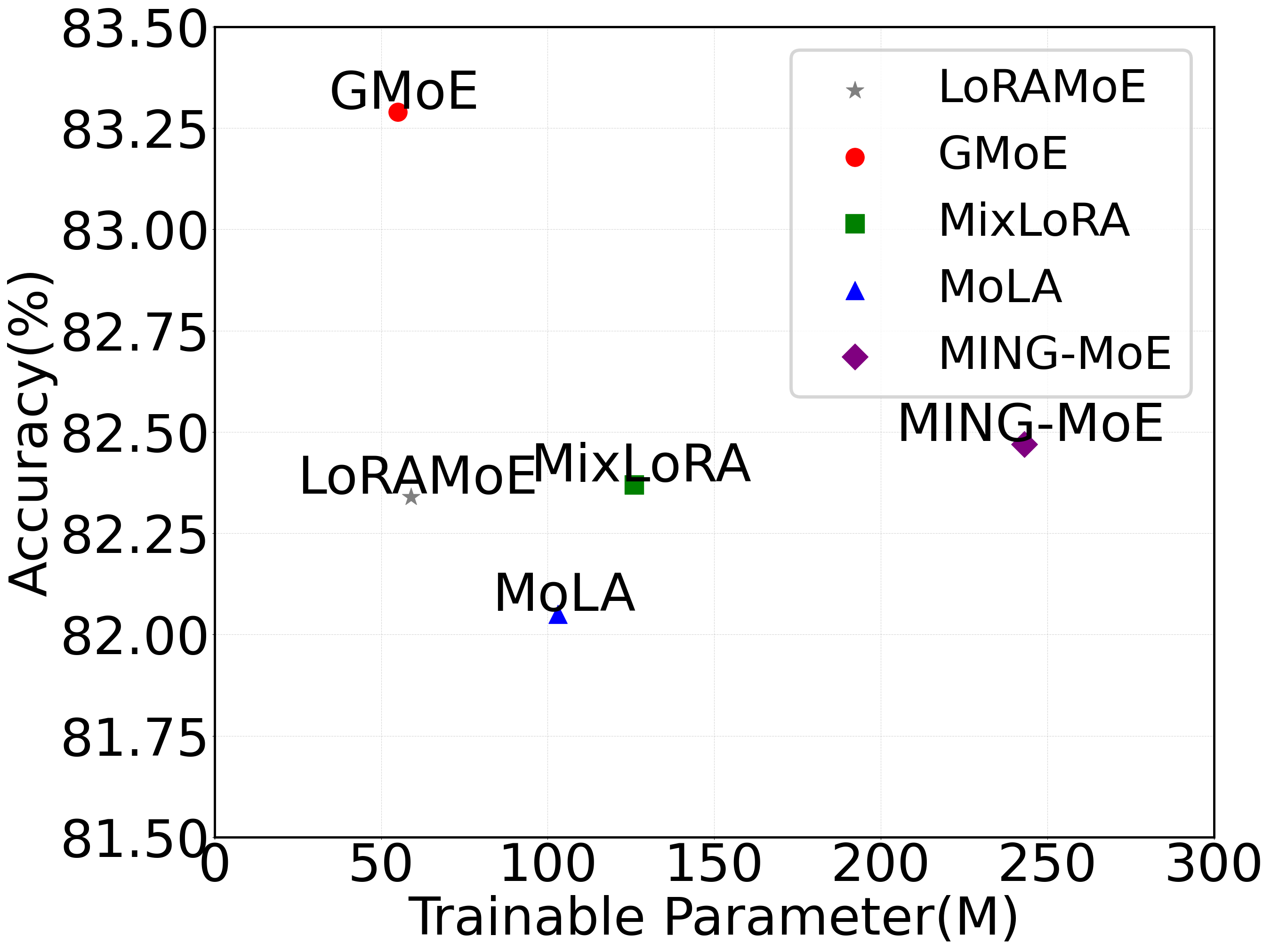}
    \caption{Qwen2-7B}
    \label{fig:sca2}
\end{subfigure}
\hfill
\begin{subfigure}[h]{0.24\linewidth}
    \centering
    \includegraphics[width=\linewidth]{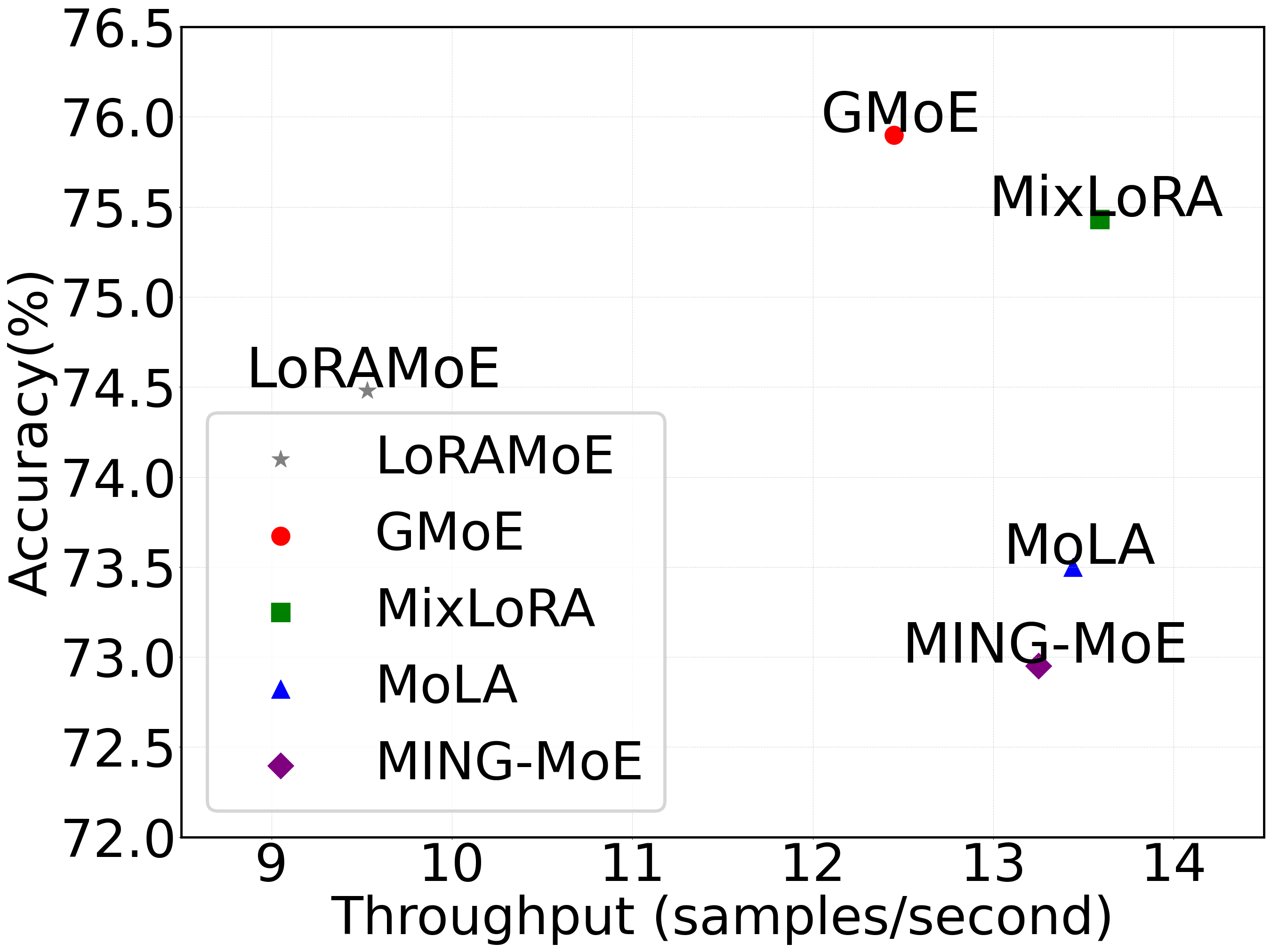}
    \caption{Llama3-8B}
    \label{fig:sca3}
\end{subfigure}
\hfill
\begin{subfigure}[h]{0.24\linewidth}
    \centering
    \includegraphics[width=\linewidth]{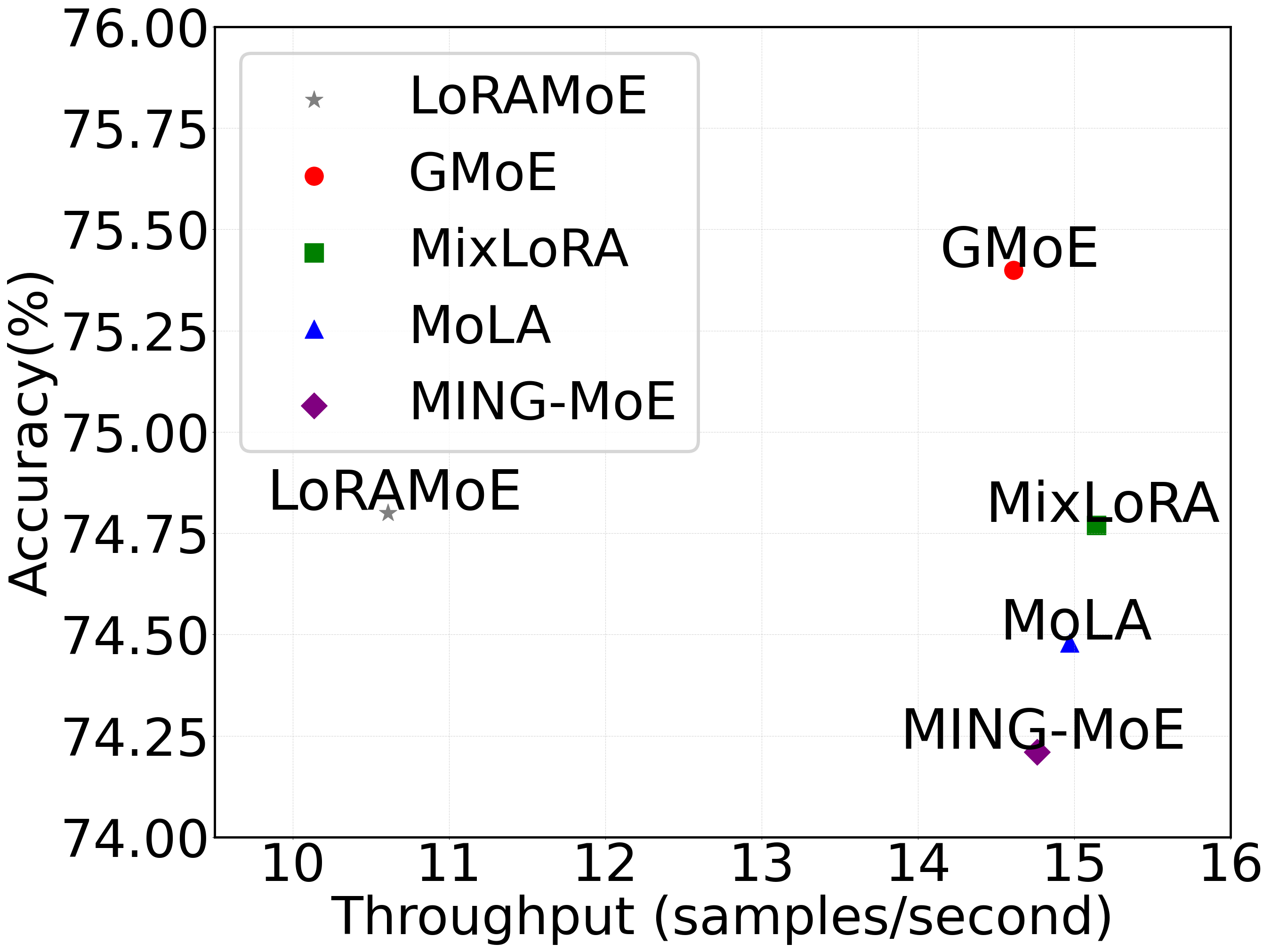}
    \caption{Qwen2-7B}
    \label{fig:sca4}
\end{subfigure}
\caption{ Efficiency comparisons across MoE methods. Illustrated with Trainable parameters and Throughput (samples/second) on the x-axis and average accuracy across four datasets on the y-axis.}
%Figure(d)'s x-axis is the Throughput based on Llama3-8B with an A800-80G}
\label{fig:scatter}
\end{figure*}

\subsection{Ablation Studies}
%  不同的损失函数模块起到的作用，可以用表格或图呈现
We perform ablation studies to demonstrate the effectiveness of the graph router and the two coordination strategies used in GMoE. 
% The graph router in GMoE allocates weight for each expert based on the collaboration information learned from the MoE graph. Besides, the Poisson distribution-based expert distinction strategy and normal distribution-based load balance strategy are adopted to empower the capability of each expert and their collaborations. 
To verify the utility of each component, we individually remove the graph router, Poisson distinct loss (see in Eq.~\ref{poisson}), and Normal balance loss (see in Eq.~\ref{norm_loss}). 
The performance of the degraded variants, namely (-Graph), (-Normal), and (-Poisson), across four datasets is illustrated in Figure~\ref{fig:abli}.
% The performance comparisons of the degradation variants (-Graph), (-Normal), (-Poisson) on four datasets are shown in Fig.~\ref{fig:abli}.
We can see that all three components make contributions to the model's performance. The degradation impact of removing the graph router (-Graph) is most obvious in ARC-Challenge and SIQA datasets in both model accuracy and stability aspects, indicating the importance of using the MoE graph to enhance expert collaborations.
Moreover, the removal of the Poisson distribution loss (-Poisson) results in the most pronounced decrease in model performance on the BoolQ and OpenBookQA datasets, showing the essential role of maintaining the distinct capability of each expert. The normal distribution loss contributes to keeping the load balance of all experts and substantially improves the model performance.

%\begin{figure}[htbp]

\subsection{Hyper-Parameter Analysis}
The performance of GMoE is affected by many hyper-parameters. We conduct analysis experiments to show the effects of hyper-parameters and 
% including the number of experts used in MoE, the number of Top-K experts, the rank in the LoRA component, the edge density, and the effect of edge connection in MoE graph construction.   
% We 
present the results of ARC-Challenge and BoolQ datasets on Qwen2-7B in Fig.~\ref{fig:para}. 
% We have the following observations.

\paragraph{The number of experts.} The number of experts varies from $\{4,8,12,16,32\}$. We can see that our model achieves the best performance with 8 experts. The increase in the number of experts may potentially complicate the collaboration process of experts and enlarge the imbalance problem. 

\paragraph{Top-K experts.} The number of experts activated by the router function with the Top-K largest assigned weights. The $K$ varies from $\{1,2,3,4,5\}$ in our experiments. We find our model performs the best with $K=2$.  

\paragraph{Rank of LoRA.} The hyper-parameter rank $r$ in LoRA controls the number of parameters that are updated in the fine-tuning process. The rank $r$ varies from $\{1,2,4,8,16,32\}$. Our model with $r=2$ achieves the best model performance which is less than the rank $r=4$ and $r=8$ required in MixLoRA in two datasets.   

\paragraph{Edge density.} The edge density controls the connection among expert nodes. The edge density is computed as the ratio of the number of edges in the MoE graph compared to the edges in the fully connected graph.  The edge density $\beta$ varies from $\{10\%,20\%,40\%,60\%,80\%\}$. We find that with only $10\%$ connected edges, the collaboration information shared among experts facilitates our model to achieve the best results.

\paragraph{Initial edge connection.} In our experiments, the edges between all expert nodes are randomly constructed. To investigate whether the edge connections among experts in the initial graph affect the model's performance, we repeat experiments under the same edge density $10\%$  with five random edge-connection settings. We can see that the initial connections among expert nodes have minimal influence on the performance of our model. Initially, all expert nodes are considered equal to obtain the input information. As training progresses, each expert node is automatically assigned corresponding weights based on the learning tasks to achieve optimal allocation effects.

\subsubsection{Efficiency Analysis} 
As shown in Fig.~\ref{fig:scatter}, we quantify architectural efficiency through the number of \textbf{Trainable parameters} and \textbf{Throughput (samples/second)}. Trainable parameters can be used as a critical metric for memory and storage costs, and Throughput can serve as a metric for computational speed that reflects the data processing rate during inference stages. These two metrics reflect the computational efficiency of the training and inference stages of our method.
We can observe that our GMoE achieves the best model performance with the least number of trainable parameters.
This phenomenon is primarily attributed to the much lower rank of LoRA required in our framework. The effective expert cooperation mechanism in our graph router reduces the number of parameters required by each expert, enabling our method to achieve both effectiveness and efficiency. In terms of model throughput, although our model does not have the highest throughput, it is comparable to other models while achieving the best model performance.

\subsection{Conclusion}

We propose GMoE, a novel framework to address the instability of LLMs caused by MoE load imbalance. GMoE enhances expert collaboration in parameter-efficient fine-tuning through effective information sharing on graph neural networks. GMoE 
introduces a novel Poisson-based expert distinction strategy to promote expert specialization while employing a normal-based load balance strategy to regulate workload distribution.
% introduces two novel strategies: Poisson-based expert distinction and normal-based load balance, 
% These two strategies comprehensively boost the capability of experts and coordinate the collaboration among them. 
The comprehensive collaboration among our sparse graph MoE architecture provides a solution to make trade-offs among the model performance, stability, and resource overhead. 

\subsection{Limitation}
% 补充limitation
We propose GMoE, a graph-based Mixture-of-Experts framework that addresses load imbalance and instability in LLM fine-tuning by enabling expert collaboration via a GNN-powered graph router.
Despite its advancements, GMoE has several notable limitations: (1) due to computational resource constraints, the framework’s effectiveness and multi-expert balance mechanism have only been validated in downstream task fine-tuning for LLMs with under 10B parameters, leaving their applicability to larger-scale pre-training settings (e.g., 100B+ parameters) untested;
(2) while initial random expert connections in the MoE Graph show minimal impact on performance, the current static graph topology does not adaptively optimize expert interactions, potentially limiting the full realization of collaborative potential and requiring future exploration of dynamic structure learning.
These gaps highlight opportunities for future research to scale GMoE to larger models, dynamic graph architectures, and expanded application domains.
% 在预训练模型上验证

\bibliography{custom}

@article{moe20222,
  title={St-moe: Designing stable and transferable sparse expert models},
  author={Zoph, Barret and Bello, Irwan and Kumar, Sameer and Du, Nan and Huang, Yanping and Dean, Jeff and Shazeer, Noam and Fedus, William},
  journal={arXiv preprint arXiv:2202.08906},
  year={2022}
}

@article{switch-trans,
  title={Switch transformers: Scaling to trillion parameter models with simple and efficient sparsity},
  author={Fedus, William and Zoph, Barret and Shazeer, Noam},
  journal={Journal of Machine Learning Research},
  volume={23},
  number={120},
  pages={1--39},
  year={2022}
}

@inproceedings{loadbal,
  title={Outrageously Large Neural Networks: The Sparsely-Gated Mixture-of-Experts Layer},
  author={Shazeer, Noam and Mirhoseini, Azalia and Maziarz, Krzysztof and Davis, Andy and Le, Quoc and Hinton, Geoffrey and Dean, Jeff},
  booktitle={International Conference on Learning Representations},
  year={2016}
}

@article{loramoe2,
  title={Mixture of cluster-conditional lora experts for vision-language instruction tuning},
  author={Gou, Yunhao and Liu, Zhili and Chen, Kai and Hong, Lanqing and Xu, Hang and Li, Aoxue and Yeung, Dit-Yan and Kwok, James T and Zhang, Yu},
  journal={arXiv preprint arXiv:2312.12379},
  year={2023}
}

@article{loramoe3,
  title={MOELoRA: An MOE-based Parameter Efficient Fine-Tuning Method for Multi-task Medical Applications},
  author={Liu, Qidong and Wu, Xian and Zhao, Xiangyu and Zhu, Yuanshao and Xu, Derong and Tian, Feng and Zheng, Yefeng},
  journal={CoRR},
  year={2023}
}

@article{loramoe4,
  title={Loramoe: Revolutionizing mixture of experts for maintaining world knowledge in language model alignment},
  author={Dou, Shihan and Zhou, Enyu and Liu, Yan and Gao, Songyang and Zhao, Jun and Shen, Wei and Zhou, Yuhao and Xi, Zhiheng and Wang, Xiao and Fan, Xiaoran and others},
  journal={arXiv preprint arXiv:2312.09979},
  year={2023}
}

@article{li2024mixlora,
  title={MixLoRA: Enhancing Large Language Models Fine-Tuning with LoRA based Mixture of Experts},
  author={Li, Dengchun and Ma, Yingzi and Wang, Naizheng and Cheng, Zhiyuan and Duan, Lei and Zuo, Jie and Yang, Cal and Tang, Mingjie},
  journal={arXiv preprint arXiv:2404.15159},
  year={2024}
}

@article{moeloracontra,
  title={Moelora: Contrastive learning guided mixture of experts on parameter-efficient fine-tuning for large language models},
  author={Luo, Tongxu and Lei, Jiahe and Lei, Fangyu and Liu, Weihao and He, Shizhu and Zhao, Jun and Liu, Kang},
  journal={arXiv preprint arXiv:2402.12851},
  year={2024}
}

@inproceedings{zadouri2023pushing,
  title={Pushing Mixture of Experts to the Limit: Extremely Parameter Efficient MoE for Instruction Tuning},
  author={Zadouri, Ted and {\"U}st{\"u}n, Ahmet and Ahmadian, Arash and Ermis, Beyza and Locatelli, Acyr and Hooker, Sara},
  booktitle={The Twelfth International Conference on Learning Representations},
  year={2023}
}

@article{arc,
  title={Think you have solved question answering? try arc, the ai2 reasoning challenge},
  author={Clark, Peter and Cowhey, Isaac and Etzioni, Oren and Khot, Tushar and Sabharwal, Ashish and Schoenick, Carissa and Tafjord, Oyvind},
  journal={arXiv preprint arXiv:1803.05457},
  year={2018}
}

@inproceedings{obqa,
  title={Can a Suit of Armor Conduct Electricity? A New Dataset for Open Book Question Answering},
  author={Mihaylov, Todor and Clark, Peter and Khot, Tushar and Sabharwal, Ashish},
  booktitle={Proceedings of the 2018 Conference on Empirical Methods in Natural Language Processing},
  pages={2381--2391},
  year={2018}
}

@inproceedings{siqa,
  title={Social IQa: Commonsense Reasoning about Social Interactions},
  author={Sap, Maarten and Rashkin, Hannah and Chen, Derek and Le Bras, Ronan and Choi, Yejin},
  booktitle={Proceedings of the 2019 Conference on Empirical Methods in Natural Language Processing and the 9th International Joint Conference on Natural Language Processing (EMNLP-IJCNLP)},
  pages={4463--4473},
  year={2019}
}

@inproceedings{boolq,
  title={BoolQ: Exploring the Surprising Difficulty of Natural Yes/No Questions},
  author={Clark, Christopher and Lee, Kenton and Chang, Ming-Wei and Kwiatkowski, Tom and Collins, Michael and Toutanova, Kristina},
  booktitle={Proceedings of NAACL-HLT},
  pages={2924--2936},
  year={2019}
}

@article{qwen2,
      title={Qwen2 Technical Report}, 
      author={An Yang and Baosong Yang and Binyuan Hui and Bo Zheng and Bowen Yu and Chang Zhou and Chengpeng Li and Chengyuan Li and Dayiheng Liu and Fei Huang and Guanting Dong and Haoran Wei and Huan Lin and Jialong Tang and Jialin Wang and Jian Yang and Jianhong Tu and Jianwei Zhang and Jianxin Ma and Jin Xu and Jingren Zhou and Jinze Bai and Jinzheng He and Junyang Lin and Kai Dang and Keming Lu and Keqin Chen and Kexin Yang and Mei Li and Mingfeng Xue and Na Ni and Pei Zhang and Peng Wang and Ru Peng and Rui Men and Ruize Gao and Runji Lin and Shijie Wang and Shuai Bai and Sinan Tan and Tianhang Zhu and Tianhao Li and Tianyu Liu and Wenbin Ge and Xiaodong Deng and Xiaohuan Zhou and Xingzhang Ren and Xinyu Zhang and Xipin Wei and Xuancheng Ren and Yang Fan and Yang Yao and Yichang Zhang and Yu Wan and Yunfei Chu and Yuqiong Liu and Zeyu Cui and Zhenru Zhang and Zhihao Fan},
      journal={arXiv preprint arXiv:2407.10671},
      year={2024}
}

@misc{llama3,
  title={Llama 3 Model},
  author={AI@Meta},
  year={2024},
  url = {https://github.com/meta-llama/llama3/blob/main/README.md}
}

@article{yi,
  title={Yi: Open foundation models by 01. ai},
  author={Young, Alex and Chen, Bei and Li, Chao and Huang, Chengen and Zhang, Ge and Zhang, Guanwei and Li, Heng and Zhu, Jiangcheng and Chen, Jianqun and Chang, Jing and others},
  journal={arXiv preprint arXiv:2403.04652},
  year={2024}
}

@inproceedings{lora,
  title={LoRA: Low-Rank Adaptation of Large Language Models},
  author={Hu, Edward J and Wallis, Phillip and Allen-Zhu, Zeyuan and Li, Yuanzhi and Wang, Shean and Wang, Lu and Chen, Weizhu and others},
  booktitle={International Conference on Learning Representations},
  year={2021}
}

@article{loramoe,
  title={Loramoe: Revolutionizing mixture of experts for maintaining world knowledge in language model alignment},
  author={Dou, Shihan and Zhou, Enyu and Liu, Yan and Gao, Songyang and Zhao, Jun and Shen, Wei and Zhou, Yuhao and Xi, Zhiheng and Wang, Xiao and Fan, Xiaoran and others},
  journal={arXiv preprint arXiv:2312.09979},
  volume={4},
  number={7},
  year={2023}
}

@article{liao2024ming,
  title={MING-MOE: Enhancing Medical Multi-Task Learning in Large Language Models with Sparse Mixture of Low-Rank Adapter Experts},
  author={Liao, Yusheng and Jiang, Shuyang and Wang, Yu and Wang, Yanfeng},
  journal={arXiv preprint arXiv:2404.09027},
  year={2024}
}

@article{mola,
  title={Higher layers need more lora experts},
  author={Gao, Chongyang and Chen, Kezhen and Rao, Jinmeng and Sun, Baochen and Liu, Ruibo and Peng, Daiyi and Zhang, Yawen and Guo, Xiaoyuan and Yang, Jie and Subrahmanian, VS},
  journal={arXiv preprint arXiv:2402.08562},
  year={2024}
}

@article{chen2024llava,
  title={Llava-mole: Sparse mixture of lora experts for mitigating data conflicts in instruction finetuning mllms},
  author={Chen, Shaoxiang and Jie, Zequn and Ma, Lin},
  journal={arXiv preprint arXiv:2401.16160},
  year={2024}
}

@article{lin2024moe,
  title={Moe-llava: Mixture of experts for large vision-language models},
  author={Lin, Bin and Tang, Zhenyu and Ye, Yang and Cui, Jiaxi and Zhu, Bin and Jin, Peng and Zhang, Junwu and Ning, Munan and Yuan, Li},
  journal={arXiv preprint arXiv:2401.15947},
  year={2024}
}

@article{li2024uni,
  title={Uni-MoE: Scaling Unified Multimodal LLMs with Mixture of Experts},
  author={Li, Yunxin and Jiang, Shenyuan and Hu, Baotian and Wang, Longyue and Zhong, Wanqi and Luo, Wenhan and Ma, Lin and Zhang, Min},
  journal={arXiv preprint arXiv:2405.11273},
  year={2024}
}

@inproceedings{houlsby2019parameter,
  title={Parameter-efficient transfer learning for NLP},
  author={Houlsby, Neil and Giurgiu, Andrei and Jastrzebski, Stanislaw and Morrone, Bruna and De Laroussilhe, Quentin and Gesmundo, Andrea and Attariyan, Mona and Gelly, Sylvain},
  booktitle={International conference on machine learning},
  pages={2790--2799},
  year={2019},
  organization={PMLR}
}

@inproceedings{li2021prefix,
  title={Prefix-Tuning: Optimizing Continuous Prompts for Generation},
  author={Li, Xiang Lisa and Liang, Percy},
  booktitle={Proceedings of the 59th Annual Meeting of the Association for Computational Linguistics and the 11th International Joint Conference on Natural Language Processing (Volume 1: Long Papers)},
  pages={4582--4597},
  year={2021}
}

@inproceedings{lester2021power,
  title={The Power of Scale for Parameter-Efficient Prompt Tuning},
  author={Lester, Brian and Al-Rfou, Rami and Constant, Noah},
  booktitle={Proceedings of the 2021 Conference on Empirical Methods in Natural Language Processing},
  pages={3045--3059},
  year={2021}
}

@inproceedings{wu2024mixture,
  title={Mixture of LoRA Experts},
  author={Wu, Xun and Huang, Shaohan and Wei, Furu},
  booktitle={The Twelfth International Conference on Learning Representations},
  year={2024}
}

@article{wang2022adamix,
  title={Adamix: Mixture-of-adapter for parameter-efficient tuning of large language models},
  author={Wang, Yaqing and Mukherjee, Subhabrata and Liu, Xiaodong and Gao, Jing and Awadallah, Ahmed Hassan and Gao, Jianfeng},
  journal={arXiv preprint arXiv:2205.12410},
  volume={1},
  number={2},
  pages={4},
  year={2022}
}

@article{zhu2023sira,
  title={Sira: Sparse mixture of low rank adaptation},
  author={Zhu, Yun and Wichers, Nevan and Lin, Chu-Cheng and Wang, Xinyi and Chen, Tianlong and Shu, Lei and Lu, Han and Liu, Canoee and Luo, Liangchen and Chen, Jindong and others},
  journal={arXiv preprint arXiv:2311.09179},
  year={2023}
}

@article{li2024locmoe,
  title={Locmoe: A low-overhead moe for large language model training},
  author={Li, Jing and Sun, Zhijie and He, Xuan and Zeng, Li and Lin, Yi and Li, Entong and Zheng, Binfan and Zhao, Rongqian and Chen, Xin},
  journal={arXiv preprint arXiv:2401.13920},
  year={2024}
}

@article{bengio2015conditional,
  title={Conditional computation in neural networks for faster models},
  author={Bengio, Emmanuel and Bacon, Pierre-Luc and Pineau, Joelle and Precup, Doina},
  journal={arXiv preprint arXiv:1511.06297},
  year={2015}
}

@article{tang2025graphmoe,
  title={GRAPHMOE: Amplifying Cognitive Depth of Mixture-of-Experts Network via Introducing Self-Rethinking Mechanism},
  author={Tang, Chen and Lv, Bo and Zheng, Zifan and Yang, Bohao and Zhao, Kun and Liao, Ning and Wang, Xiaoxing and Xiong, Feiyu and Li, Zhiyu and Liu, Nayu and others},
  journal={arXiv preprint arXiv:2501.07890},
  year={2025}
}

@article{tian2024hydralora,
  title={Hydralora: An asymmetric lora architecture for efficient fine-tuning},
  author={Tian, Chunlin and Shi, Zhan and Guo, Zhijiang and Li, Li and Xu, Cheng-Zhong},
  journal={Advances in Neural Information Processing Systems},
  volume={37},
  pages={9565--9584},
  year={2024}
}

\end{document}